\documentclass[letterpaper]{nature}

\usepackage{color}
\usepackage{url}
\usepackage{verbatim}
\usepackage{multirow}
\usepackage{xspace}
\usepackage{graphicx}
\usepackage{amsmath} 
\usepackage{bbm}
\usepackage{amssymb}
\usepackage{booktabs} 
\usepackage{times} 
\usepackage[OT1]{fontenc}
\usepackage{lscape}
\usepackage[left=0.66in,right=0.66in,bottom=0.70in,top=0.62in,headheight=15pt,headsep=12pt]{geometry}
\usepackage{fancyhdr}
\usepackage{enumitem}
\usepackage[table,xcdraw]{xcolor}
\usepackage[normalem]{ulem}
\usepackage{xcolor}
\usepackage[innercaption]{sidecap}
\usepackage{dirtytalk}
\usepackage{newunicodechar}
\newunicodechar{’}{\textquoteright}
\newunicodechar{–}{\textendash}
\usepackage[most]{tcolorbox}
\usepackage{microtype}
\usepackage{titlesec}
\usepackage[Export]{adjustbox}
\adjustboxset*{max width=\linewidth}

\let\naturespacing\spacing
\let\spacing\relax 
\usepackage{setspace}
\let\spacing\naturespacing
\usepackage[colorlinks,citecolor=blue,urlcolor=magenta]{hyperref}
\hypersetup{pdftitle={DiscoveryAgent: A Framework for Scientist-centric Agentic AI},pdfauthor={Owen Queen et al.}}
\usepackage[margin=0cm, labelfont=bf, font=scriptsize]{caption}
\DeclareCaptionFont{figuretext}{\fontsize{10pt}{12pt}\selectfont}
\newsavebox{\figurecaptionbox}
\newlength{\figureimageheight}
\NewDocumentCommand{\preparefigurecaption}{+m m}{%
  \sbox{\figurecaptionbox}{\begin{minipage}{\linewidth}%
    \caption{#1}\label{#2}%
  \end{minipage}}%
  \setlength{\figureimageheight}{\dimexpr\textheight-\ht\figurecaptionbox-\dp\figurecaptionbox-4pt\relax}%
}
\usepackage[ruled,vlined]{algorithm2e}
\SetKwRepeat{Do}{do}{while}

\DeclareCaptionLabelFormat{extended_data}{\textbf{Extended Data #1 #2: }}

\newcommand{\xhdr}[1]{\vspace{1.7mm}\noindent{{\bf #1.}}}

\usepackage[normalem]{ulem}
\usepackage{bm}
\usepackage{amsthm}
\theoremstyle{definition}

\graphicspath{{./FIG/}}

\newcommand{\name}{\textsc{DiscoveryAgent}\xspace}

\definecolor{draftocean}{HTML}{F0F5F0}
\definecolor{draftoceanink}{HTML}{202A32}
\definecolor{draftoceanaccent}{HTML}{426755}
\definecolor{steeringorange}{HTML}{FAF5EA}
\definecolor{steeringorangeframe}{HTML}{916523}
\definecolor{steeringorangeink}{HTML}{916523}

\tcbset{takeawaybase/.style={enhanced,arc=0pt,outer arc=0pt,boxsep=0pt,
  colback=white,coltext=draftoceanink,colbacktitle=white,coltitle=draftoceanink,
  fontupper=\normalsize,fontlower=\normalfont\normalsize,collower=draftoceanink,
  fonttitle=\bfseries\fontsize{12pt}{14.5pt}\selectfont,
  left=13pt,right=13pt,top=11pt,bottom=12pt,toptitle=11pt,bottomtitle=2pt,
  before skip=18pt,after skip=18pt}}

\tcbset{takeawayframe/.style={takeawaybase,boxrule=2pt,top=15pt,
 attach boxed title to top left={xshift=12pt,yshift=-\tcboxedtitleheight/2},
 fonttitle=\fontfamily{phv}\bfseries\fontsize{10pt}{12pt}\selectfont,coltitle=white,
 boxed title style={enhanced,frame hidden,arc=0pt,boxsep=0pt,left=7pt,right=7pt,top=4pt,bottom=4pt}}}
\newtcolorbox{casestudytakeaways}[1]{takeawayframe,breakable=false,colframe=draftoceanaccent!70!white,colbacktitle=draftoceanaccent,title={#1}}
\newtcolorbox{steeringobservations}[2][]{takeawayframe,breakable,lines before break=5,colframe=steeringorangeink!70!white,colbacktitle=steeringorangeink,title={#2},extras middle and last={overlay={
 \node[anchor=west,fill=steeringorangeink,text=white,inner xsep=7pt,inner ysep=4pt,
 font=\fontfamily{phv}\bfseries\fontsize{10pt}{12pt}\selectfont]
 at ([xshift=12pt]frame.north west) {#2 (continued)};}},#1}

\titleformat{\section}
  {\sffamily\color{draftoceanink}\bfseries\fontsize{13.5pt}{15pt}\selectfont}
  {\thesection}{0.65em}{}
\titleformat{\subsection}
  {\sffamily\color{draftoceanink}\bfseries\fontsize{12pt}{14pt}\selectfont}
  {\thesubsection}{0.6em}{}
\titleformat{\subsubsection}
  {\sffamily\color{draftoceanink}\bfseries\fontsize{9.3pt}{10.4pt}\selectfont}
  {\thesubsubsection}{0.55em}{}
\titlespacing*{\section}{0pt}{10pt plus 2pt minus 2pt}{4.5pt}
\titlespacing*{\subsection}{0pt}{8pt plus 1pt minus 1pt}{3pt}
\titlespacing*{\subsubsection}{0pt}{6pt plus 1pt minus 1pt}{2pt}

\begin{document}
\color{draftoceanink}

\fontsize{9pt}{10pt}\selectfont
\spacing{1.00}
\begin{tcolorbox}[
  enhanced,
  frame hidden,
  colback=draftocean,
  boxrule=0pt,
  arc=3mm,
  left=15pt,
  right=15pt,
  top=16pt,
  bottom=16pt,
  before skip=0pt,
  after skip=13pt
]
{\raggedright\sffamily\color{draftoceanaccent}\bfseries\fontsize{23pt}{26pt}\selectfont
DiscoveryAgent: A Framework for Scientist-centric Agentic AI\par}
\vspace{10pt}
{\raggedright\sffamily\color{draftoceanink}\fontsize{11pt}{14pt}\selectfont
\mbox{Owen Queen\textsuperscript{1}},
\mbox{Ioana Ciuc\u{a}\textsuperscript{1,2,*}},
\mbox{Emma Dann\textsuperscript{1,3,*}},
\mbox{Jean Somalwar\textsuperscript{1,2,4,*}},
\mbox{Carine You\textsuperscript{1,*}},
\mbox{Steven Dillmann\textsuperscript{1,2}},
\mbox{Viraj Manwadkar\textsuperscript{1,2}},
\mbox{Evan Gorrell\textsuperscript{5}},
\mbox{Harrison G. Zhang\textsuperscript{1}},
\mbox{Jonathan K Pritchard\textsuperscript{1}},
\mbox{Risa H Wechsler\textsuperscript{1,2,6}},
\mbox{James Zou\textsuperscript{1}}\par}

\vspace{14pt}
\noindent{\sffamily\bfseries\color{draftoceanaccent}\fontsize{8pt}{10pt}\selectfont ABSTRACT\par}
\vspace{4pt}
{\color{draftoceanink}\normalsize\selectfont\noindent AI agents can assist scientific research, but balancing autonomous investigation with scientists’ ability to inspect and redirect ongoing work remains a central design challenge.
We present \name, a domain-agnostic, scientist-centric system that combines autonomous investigation with scientist-directed hypothesis steering. 
Scientists select, refine, or replace hypotheses between research rounds, while audit agents interrogate checkpointed records to trace claims, reproduce analyses, and inform subsequent decisions. 
Across two predictive modeling tasks, \name produced the best variable-star classifier among evaluated baselines and, with scientist steering, a protein variant-effect predictor surpassed by only one of 106 ProteinGym baselines.
We develop a rigorous checkpointing system that enables highly capable audit agents, a scalable form of oversight for agentic research.
Audit agents autonomously identify errors at higher rates than previous AI scientist agents, and scientists collaborating with audit agents identify 14x more valid errors versus scientists performing manual review.
In three case studies led by domain experts across physical science, social science, and life science, we developed a multimodal photometric-redshift estimator that resolves complementary failures of images and photometry, achieving 0.971 AUROC for low-redshift classification; documented an 83\% rise in Florida homeowners-insurance premiums from 2013 to 2023, with a growing, regressive burden; and identified CD4$^{+}$ T-cell relay genes, which both receive and exert broad transcriptional regulation, enriched for autoimmune disease associations. 
Together, these results show how autonomous research can expand scientists' empirical reach while centering their authority over research direction and their ability to understand and challenge the evidence.\par}
\end{tcolorbox}

\begingroup
\renewcommand{\thefootnote}{}
\makeatletter
\renewcommand{\@makefntext}[1]{\noindent#1}
\makeatother
\begin{NoHyper}
\footnotetext{%
\raggedright\fontsize{8pt}{9.5pt}\selectfont
\textsuperscript{1}Stanford University;\enspace
\textsuperscript{2}Kavli Institute for Particle Astrophysics and Cosmology (KIPAC);\enspace
\textsuperscript{3}Gladstone-UCSF Institute of Genomic Immunology;\enspace
\textsuperscript{4}University of California, Berkeley;\enspace
\textsuperscript{5}University of Tennessee, Knoxville;\enspace
\textsuperscript{6}SLAC National Accelerator Laboratory;\enspace
\textsuperscript{*}Core contributing scientists}
\end{NoHyper}
\endgroup

\setcounter{secnumdepth}{3}

\section{Introduction}
\label{sec:intro}
Scientific discovery rarely begins and ends with a single hypothesis test.
It is an iterative process in which broad questions become concrete hypotheses, experiments expose unexpected patterns, and results reshape beliefs and the questions worth asking next.
Large language model (LLM) agents can already accelerate components of this process, including scientific software development and data analysis~\cite{aygun2026aisystem,merrill2026terminalbench,nie2026dsgym, Terminal-Bench-Science_2026}, as well as literature synthesis and scientific reasoning~\cite{gottweis2026accelerating,bubeck2025early}.
Combining these capabilities has motivated ``AI scientists'' that plan and conduct increasingly complete research workflows, from ideation through experimental execution~\cite{lu2026towards,ghareeb2026multi,mitchener2025kosmos,agarwal2026autodiscovery}.
These systems promise to expand the range of ideas scientists can explore, but the defining challenge is not simply to automate more steps.
Scientific progress depends on deciding which observations are consequential, which assumptions should be challenged, and which direction merits the next experiment.
A system that scales execution while marginalizing this judgment can produce more research without producing better science.

Current agentic systems tend to occupy three distinct regions of this design space.
General-purpose coding agents are highly interactive and locally steerable: scientists can request an analysis, inspect its implementation, and redirect the next turn.
However, the unit of work is a bounded software or analysis objective, leaving the scientist responsible for decomposing a long investigation, integrating evidence across tasks, and maintaining empirical history over days to weeks.
Verifier-guided agents can autonomously search broad solution spaces when progress is captured by an executable evaluator~\cite{novikov2025alphaevolve}, but this assumption does not hold for open-ended questions whose objectives are hard to define or change as evidence accumulates.
End-to-end AI scientists sustain longer planning and experimentation loops, but they place the scientist at the beginning and end of the process.
In this setting, the scientist supplies an objective and data, the system determines what to investigate, and the resulting trajectory is compressed into a final report with no intermediate input from scientists.
Scientists are therefore left between two unsatisfactory roles---actively decomposing and supervising each analysis, or passively reviewing a large body of research in which they played little part.

We argue that a truly scientist-centric AI discovery system should instead satisfy three desiderata.
First, \textit{steerability} should operate at the level of scientific direction.
An expert should be able to select, refine, or replace hypotheses as results emerge, while autonomous agents implement the well-scoped analyses required to test them.
This division of labor concentrates valuable expert attention on decisions that depend on domain knowledge, tacit understanding, and research taste.
Previous work has shown that LLM-generated ideas appear more novel at proposal but are less feasible~\cite{si2025can}, lose novelty after execution~\cite{si2025ideation}, and exhibit substantial homogeneity across repeated generations~\cite{jiang2025artificial}.
Scientist-centric interaction should therefore prioritize autonomy where agents are strongest while keeping experts in control of what constitutes an interesting and useful research direction.

Second, \textit{transparency} requires the evolving research state to remain legible and reversible.
Every agent action, analysis, and artifact should be recorded; code and data products should be version-controlled; and checkpoints should preserve the exact state from which a finding emerged.
A final report or monolithic notebook does not by itself reveal how evidence propagated across hypotheses, which negative results redirected the investigation, or whether a later claim rests on an earlier analytical error.
For a long-horizon research system, the research history must be directly indexable and treated as a first-class output rather than a byproduct of agentic work.

Third, beyond transparency, a scientist-centric AI discovery system should exhibit \textit{auditability}, meaning that this record can be interrogated and probed for reproducibility and claim verification.
To enable post hoc research evaluation, each claim should be traceable to its supporting analyses, each result should be reproducible from artifacts, and consequential assumptions should be open to targeted challenge.
This is a necessary component as analyses of AI scientist system traces have identified selective use of evidence, conclusions unsupported by executed experiments, and failures to revise beliefs after contradictory results~\cite{luo2026more,rios2026ai}.
Beyond post hoc research evaluation, auditability enables effective steering by allowing scientists to actively probe the research during their decision-making, understanding how results and methods evolve over the project. 
Transparency makes evidence available, but details get lost if a scientist is required to manually review millions of lines of logs and code before every decision.
Auditability unlocks scalable oversight of autonomous research, where scientists can use AI as a collaborator to assist in research understanding and comprehension while performing scientific quality assurance at scale.

Here, we introduce \name, a scientist-centric AI discovery system designed around these desiderata.
\name structures research as an outer loop of hypothesis branches and an inner loop of research queries (Figure \ref{fig:overview}a).
Within each branch, a director autonomously coordinates specialized agents for literature search, data analysis, and review; across branches, empirical memory consolidates the evidence needed to choose what to investigate next, while procedural memory records how the growing research codebase can be reused.
At the boundary between branches, \textit{hypothesis steering} allows a scientist to inspect the accumulated evidence and then select, edit, or replace the next hypothesis (Figure \ref{fig:overview}b).
This design decouples scientific direction from computational supervision; agents retain autonomy to execute scoped investigations, while the scientist retains authority over how the broader inquiry develops.

To make this steering informed, \name records every query and artifact in a structured checkpoint tree, and places the research codebase, data products, and figures under version control.
The agent's outputs are fully accessible, allowing scientists to fully probe internal agentic reasoning.
We introduce \textit{audit agents}, which operate outside the research-generating stream and attach to the \name checkpoints to enable full provenance over the research process (Figure \ref{fig:overview}c).
A scientist can ask why an analysis was performed, identify the evidence behind a claim, reproduce a result, conduct a follow-up from an earlier state, or roll back the research when an assumption or implementation proves unsound.
In corrupted-claim experiments, a fixed auditor identifies synthetically inserted errors in \name traces with 96.5\% F1; in a human review study, participants with audit-agent access report findings matching inserted errors at nine times the proportion observed for participants given only the paper and code.
Audit agents also exposed an instance of test-set access during model development, allowing us to locate the responsible round, restore an earlier checkpoint, harden the sandbox, and re-run the affected research.
Auditability is not only a reporting property, but a mechanism that increases scientists' understanding and enables effective steering.

We demonstrate \name's efficacy across five settings spanning physical science, life science, and social science.
In variable-star classification, \name autonomously builds a model that achieves the highest held-out performance against expert hand-crafted and agent baselines.
With hypothesis steering for protein variant-effect prediction, \name develops a model that outperforms 99\% of 106 reported ProteinGym baselines, including other popular AI scientist models. We also show that expert-steered \name outperforms an autonomous run of \name, supporting the efficacy of scientist-centric, AI-driven research.
Across three in-depth case studies led by external domain experts (Figure \ref{fig:overview}d), \name and an astronomer build a multimodal photometric-redshift model that reaches 0.971 AUROC for identifying low-redshift galaxies and reveals how image morphology and photometry resolve complementary failure modes; \name and an economist show that homeowners-insurance premiums in Florida are growing more regressive---the lowest income quartile spent 6.6\% of income on insurance in 2023 versus 1.2\% for the highest, up from 4.9\% and 1.0\% in 2018, respectively; 
and finally, \name and a computational biologist establish an association between perturbation signatures and disease genetics by identifying relay genes, a class enriched for both autoimmune disease genome-wide association study hits and primary-immunodeficiency genes in CD4$^{+}$ T cells.

\name supports distinct modes of scientific work, from autonomous predictive modeling to expert-directed, open-ended inquiry. 
Hypothesis steering allows scientists to reshape investigations as evidence accumulates, while audit agents make the underlying analyses traceable, reproducible, and correctable. 
Together, these results show that \name can sustain long-horizon research across domains while allowing scientists to direct and understand the work, paving the way to scientist-centric AI discovery systems.

\section{Results}
\label{sec:results}
We first introduce \name, a generalized scientific discovery agent. We then demonstrate \name on a variety of discovery-centric tasks, including predictive modeling in astronomy and biology as well as three open-ended discovery tasks with expert collaborators.

\subsection{An Agentic System for Accelerating Discovery}
\label{sec:results:system}

\xhdr{\name architecture}
\name is organized into two nested loops that allow scientists to delegate implementation labor while retaining high-level control of project direction.
In the inner loop, a director agent autonomously tests a single hypothesis by issuing \textit{queries} to a toolbox of sub-agents for code execution and data analysis, literature review, web search, and domain-specific data access.
Each sub-agent returns a \textit{query report} that informs the next query, and a review agent periodically critiques the process (Methods Secs.~\ref{sec:methods:director}, \ref{sec:methods:tools}, \ref{sec:methods:review}, \ref{sec:methods:specialized}; Supplementary Note~\ref*{si:note:execution_agent}; Supplementary Figure~\ref*{si:fig:agent_innerloop}).
The outer loop scopes this autonomy into \textit{hypothesis branches}, the unit at which scientists intervene.
The hypothesis generation agent proposes a hypothesis conditioned on prior findings, the director explores it, and the resulting \textit{round report} is integrated into memory (Figure~\ref{fig:overview}a; Methods Sec.~\ref{sec:methods:outerloop}).
Memory has two modules: empirical memory, a natural-language ``memory report'' that a multi-agent system updates each round to accumulate relevant findings while pruning redundancy, and procedural memory, which tracks the codebase for the execution agent (Methods Sec.~\ref{sec:methods:memory}; Supplementary Notes~\ref*{si:note:empirical_memory}, \ref*{si:note:procedural_memory}).

\begin{figure}[!tp]
    \centering
    \preparefigurecaption{\textbf{\name supports scientist-steered, auditable discovery across domains.}
    \textbf{a)} Starting from raw data, \name conducts iterative research rounds and accumulates findings through empirical memory updates; between rounds, hypothesis steering determines the next direction while the codebase and all internal reasoning persist across the investigation.
    \textbf{b)} During hypothesis steering, the scientist uses a hypothesis-generation agent to develop candidate directions and an audit agent to interrogate the complete research trace before selecting the final hypothesis for the next round.
    \textbf{c)} The entire research is organized into individual queries, with original queries, query reports from the results of query execution, and codebase snapshots (if applicable) in each query. This allows scientists and audit agents to inspect individual queries, query reports, and the corresponding codebase.
    \textbf{d)} We validate \name in three case studies with experts across extragalactic astronomy, economics, and functional genomics: multimodal photometric-redshift prediction for low-redshift galaxies, a regressive and worsening home-insurance burden in hurricane-exposed markets, and enrichment of autoimmune-disease genetic drivers among ``relay genes'' in CD4$^{+}$ T cells. Shown are data representations at the top of each case study and real data from the case study at the bottom. See later figures for full explanations.
    }{fig:overview}%
    \includegraphics[width=\linewidth,height=\figureimageheight,keepaspectratio]{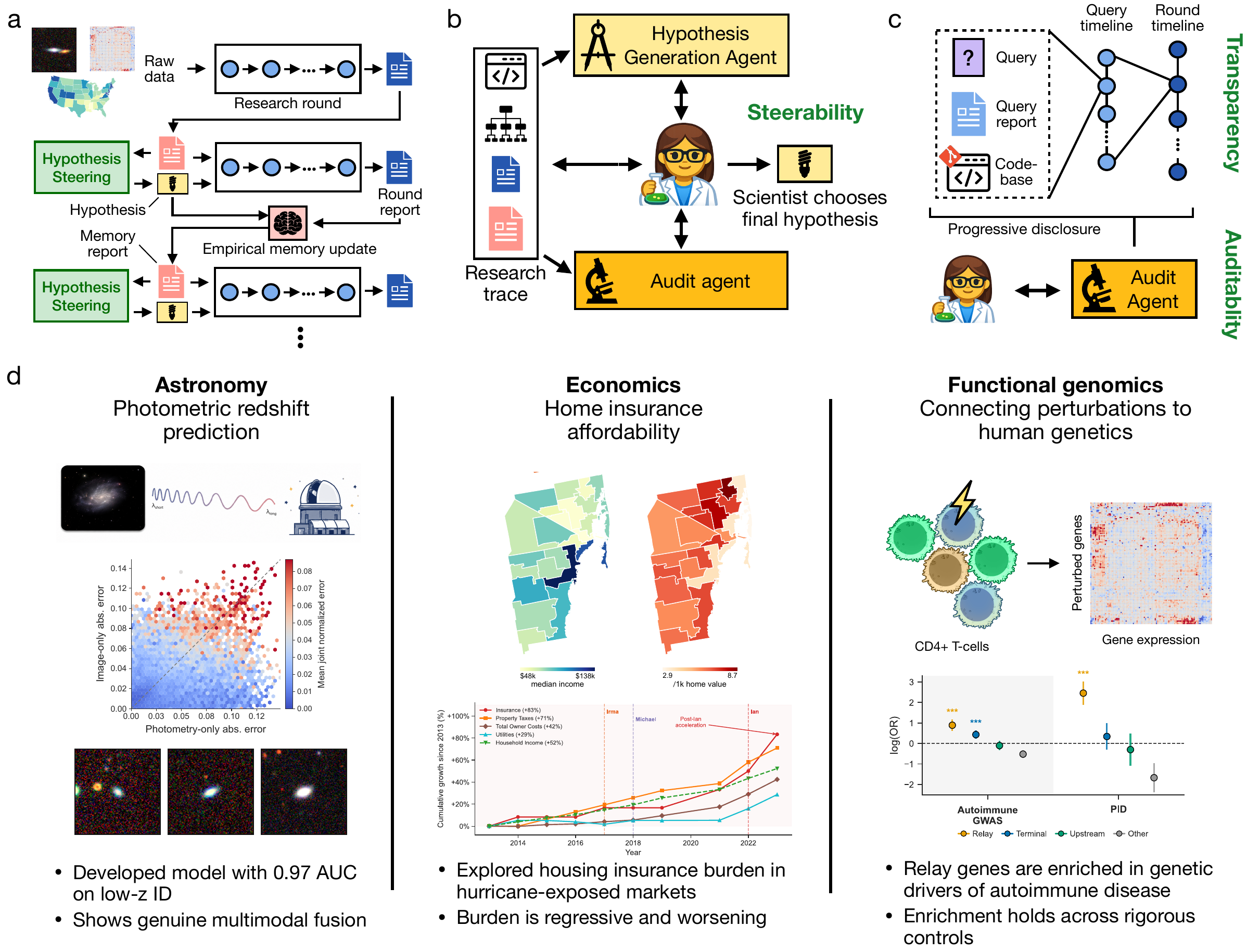}%
    \par\nointerlineskip\usebox{\figurecaptionbox}
\end{figure}

\xhdr{Hypothesis steering}
The selection of hypotheses for \name is a critical component, requiring intentional, well-informed scientific judgement.
LLMs have shown successful performance in well-scoped scientific coding tasks and literature review but with more limited success in hypothesis generation~\cite{si2025can, si2025ideation}.
Thus, we posit that human scientist direction in hypothesis generation is a high-leverage step for increasing the quality of the research process.
We introduce \textit{hypothesis steering}, a framework in which scientists drive the research process at the hypothesis selection stage (Figure \ref{fig:overview}b, Methods Sec.~\ref{sec:methods:hitl}).
During hypothesis formulation, we use a hypothesis generation agent, which views the accumulated memory report and generates candidate hypotheses.
The scientist then goes through an iterative process of editing, changing, or regenerating these hypotheses until selecting one for the next round of investigation.
During this phase, the scientist is also given access to the audit agent in order to better understand the research that has been conducted.
The audit agent is used to better inform the hypothesis selection and keep the scientist as an active participant in the research process.
Our editing and selection process is implemented through a user interface (Supplementary Note~\ref*{si:note:hypothesis_steering}).
\name can also operate without the scientist-in-the-loop.
In the ``autonomous'' mode, \name is allowed to select its own hypotheses, pursuing the hypothesis proposed by the hypothesis generation agent without scientist input.
Hypothesis steering is a central component of the \name design and enables the scientist to have a central, flexible role as the driver of the research.

\xhdr{Transparency} Maintaining rigor in autonomous research requires high transparency into agentic work. \name checkpoints every action, from code execution to literature search to internal ideation, yielding strict version control over agent-generated code and research from which the full process can be reconstructed (Methods Sec.~\ref{sec:methods:provenance}).
Further, the memory report is written in natural language, allowing the scientist to interrogate and understand the research process.

\xhdr{Audit agents}
Managing a long autonomous research project requires the scientist to understand what the agent has done.
However, requiring scientists to manually review tens of thousands of lines of autonomously generated code to detect errors defeats the purpose of high-level direction of the research.
To combat this, we introduce \textit{audit agents}, which attach to the checkpoints, codebase, and logs of \name and let the scientist interrogate the research process (Figure~\ref{fig:overview}c; Methods Sec.~\ref{sec:methods:audit_agents}).
Empowered by full access to these artifacts, audit agents let scientists ask questions about the research process, verify intermediate claims, or run follow-up analyses directly on top of \name's progress, shifting the scientist's role from reviewing raw output to directing the investigation.
This access enables more informed oversight of the research project, ultimately improving hypothesis steering and making it more effective for challenging, real-world research projects.

\xhdr{Contrasting \name to alternative agentic systems}
\name differs from prior agentic systems along three complementary axes: steerability, transparency, and auditability.
General-purpose coding agents such as Claude Code~\cite{anthropic2025claudecode} and Codex~\cite{openai2025codexsystemcard} are interactive and locally steerable, but their unit of work is a user-specified software task, so the scientist must pose each successive research question, integrate evidence across tasks, and reconstruct a research record from transcripts, diffs, and outputs that were never organized as one.
Verifier-based agents such as AlphaEvolve and AutoResearch-style systems remove the scientist entirely, autonomously searching against a strong verifier that most scientific tasks do not have.
End-to-end AI scientists such as Kosmos and AutoDiscovery are closest to \name in sustaining multi-step hypothesis generation and analysis, but they also place the scientist outside the loop, receiving discoveries only after the run.
Further, their reproducibility artifacts often bury agentic reasoning in large notebooks or unstructured logs, a lack of transparency that makes comprehension of the research process challenging.
\name instead preserves autonomy within well-scoped hypothesis branches while placing interaction at hypothesis boundaries, where audit agents interrogate the complete checkpointed trace and the scientist selects, revises, or replaces the next hypothesis.
This enables a scientist to manage the AI as a collaborator across a practical, months-scale research project rather than supervise it turn by turn or receive its output at the end.

\subsection{Predictive modeling with \name}
\label{sec:results:predictive}

While \name is designed for open-ended exploration tasks, we first test \name on two frontier tasks in time-domain astronomy and protein variant effect prediction (VEP).
Predictive tasks are a useful domain for comparison as test set metrics allow for direct comparison between the products (models) built by different agents.
We evaluate these models on test sets withheld from the agents during research (Methods Sec.~\ref{sec:methods:predictive_evaluation}) in order to test generalization.
These experiments ask whether \name can develop competitive predictors autonomously and whether domain-expert steering can further improve the resulting models.
Shared model assignments, operating modes, and research budgets are reported in Methods Sec.~\ref{sec:methods:configurations}.

\xhdr{Classification of variable stars}
We first evaluate the autonomous setting of \name on a classification task in observational time-domain astronomy.
The StarEmbed dataset \cite{li2025starembed} provides $\sim$40,000 expert-labeled, multi-band light curves from the Zwicky Transient Facility \cite{masci2019zwicky}, spanning seven classes of periodic variable stars (Figure~\ref{fig:predictive}a).
Using the same data and experimental setup as in \cite{li2025starembed}, we evaluate classifiers that predict variable star type from raw light curves and catalog periods.
We compare against six baselines: three autonomous AI-scientist systems (Kosmos \cite{mitchener2025kosmos}, AutoDiscovery \cite{agarwal2026autodiscovery}, and AI-Scientist-v2 \cite{lu2026towards}), a coding agent given a one-shot prompt (Coding agent@1), a coding agent with a fixed-goal continuation loop (Coding agent loop), and ML models trained on a curated set of expert hand-crafted features (HCF) from the original StarEmbed paper.
We use balanced accuracy to account for severe class imbalance (Methods Sec.~\ref{sec:methods:baselines}; Supplementary Note~\ref*{si:note:starembed_agent_baselines}).

On the common test set, \name's model achieves the highest balanced accuracy (86.23\%), followed by Kosmos (85.12\%) and Coding agent loop (84.05\%; Figure~\ref{fig:predictive}b).
These three agents exceed the expert HCF model (83.56\%), whereas AutoDiscovery and Coding agent@1 achieve 83.09\% and 80.91\%, respectively.
AI-Scientist-v2's model scores the lowest at 63.43\% (Supplementary Note~\ref*{si:note:starembed_classifier}).
\name's 2.67-point margin over HCF and 2.18-point margin over Coding agent loop are significant ($p=0.0049$ and $p=0.0040$, respectively), although its 1.11-point margin over Kosmos is not ($p=0.194$, Supplementary Table~\ref*{si:tab:starembed_paired}).
These results support \name's competitiveness as an autonomous research agent, a validation of the agent's internal design, even without expert steering.

The strongest agent-derived classifiers combine phase-resolved light-curve features with boosted trees.
\name's final model is a simple, flat seven-class LightGBM classifier over 46 features, including per-band distributional statistics, three-harmonic Fourier descriptions of phase-folded light curves, log-period, and phase-resolved color information (Supplementary Note~\ref*{si:note:starembed_da_model}).
Kosmos also uses LightGBM, but with 175 features combining summary statistics, Fourier morphology, and phase-bin measurements; Coding agent loop averages XGBoost and LightGBM predictions over 108 features, including wavelet, Fourier, and autocorrelation descriptors.
However, \name's results showed that more features did not necessarily translate to better performance on this benchmark.
In comparison, Coding agent@1 uses a Random Forest over 53 predominantly summary features, while AutoDiscovery's evaluated XGBoost uses only ten period-and-summary-statistic features.
Thus, \name concentrates light-curve shape and cross-band information into fewer dimensions than the two strongest agent comparators and the 138-feature expert HCF representation (Supplementary Tables~\ref*{si:tab:starembed_balacc} and~\ref*{tab:features}).

\begin{figure}[!tp]
    \centering
    \preparefigurecaption{\textbf{\name develops competitive predictive models autonomously and with scientist steering.}
    \textbf{a)} Diagram describing variable-star classification from light curves and catalog periods in StarEmbed.
    \textbf{b)} Balanced accuracy of the selected classifiers on the common StarEmbed test subset. Error bars show $\pm1$ bootstrap standard error from 2,000 row resamples. HCF denotes expert hand-crafted features.
    \textbf{c)} Protein variant effect prediction using sequence, structure, and mutation information. The heatmap depicts measured deep mutational scanning (DMS) effects in a region of DNJA1; the scatter plot compares DA-Steer predictions with measured scores.
    \textbf{d)} Left: 113 predictors ranked by median per-assay Spearman correlation ($\rho$) on ProteinGym's contiguous-position holdout; horizontal bars show the interquartile range across assays. Right: the top ten predictors, with points showing individual assays, boxes showing medians and interquartile ranges, and whiskers spanning the 10th--90th percentiles. DA-Auto denotes autonomous \name and DA-Steer denotes scientist-steered \name. Kosmos is excluded because its run retrieved benchmark labels and scores. DA-Steer exceeds DA-Auto and both coding-agent-loop variants in paired two-sided Wilcoxon signed-rank tests (Supplementary Table~\ref*{si:tab:dms_paired}).
    }{fig:predictive}%
    \includegraphics[pagebox=cropbox,clip,width=\linewidth,height=\figureimageheight,keepaspectratio]{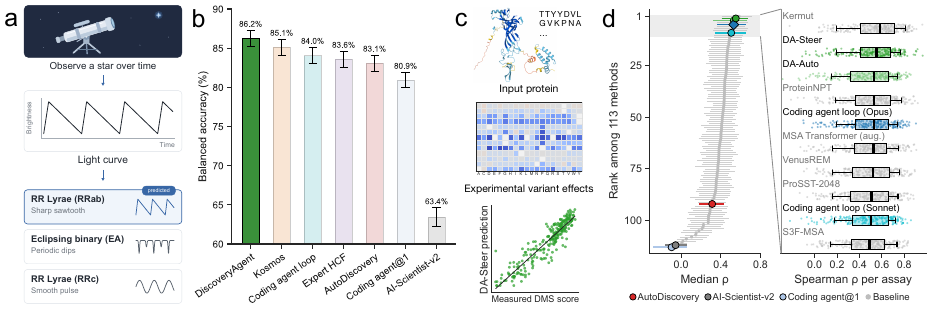}%
    \par\nointerlineskip\usebox{\figurecaptionbox}
\end{figure}

\xhdr{Predicting protein variant effects}
We next apply \name to protein variant effect prediction (VEP), predicting how amino acid substitutions alter protein function using ProteinGym's 217 deep mutational scanning (DMS) assays across 47 species~\cite{notin2023proteingym}.
DMS systematically measures the functional effects of mutations, mapping protein fitness landscapes that capture effects on phenotypes ranging from protein expression and biochemical activity to organismal fitness~\cite{fowler2014deep} (Figure~\ref{fig:predictive}c).
Accurate prediction of DMS readouts can distinguish pathogenic mutations from benign variation, identifying molecular drivers of disease and pinpointing therapeutic targets.
We compare fully autonomous DA-Auto, an autonomous instance of \name, and DA-Steer, a version guided by a machine learning scientist experienced in protein modeling, against the same baseline agents used in StarEmbed; Kosmos is excluded after it was found to have retrieved benchmark labels and scores from the internet during execution~(Supplementary Note~\ref*{si:note:kosmos_exclusion}).
We run both Sonnet 4.6 (Sonnet) and Opus 4.6 (Opus) bases for the coding agent loops.
We evaluate performance using median Spearman correlation across assays on ProteinGym's contiguous holdout split (Supplementary Note~\ref*{si:note:baseline_agents_dms} and Supplementary Table~\ref*{si:tab:dms_results}).

Guided by the scientist toward directly modeling mutation effects, DA-Steer uses a supervised 98-feature LightGBM model to learn nonlinear interactions among ESM-2 scores, physicochemical changes, multiple-sequence alignment (MSA) features, and structural features from mutation-level DMS labels (Supplementary Note~\ref*{si:note:dms_steer_architecture}).
Its structural representation includes a three-layer graph attention network (GAT)~\cite{velickovic2018graph} on residue-level contact graphs, whose lightweight architecture enabled broad hyperparameter exploration.
DA-Auto instead combines pretrained zero-shot sequence- and structure-conditioned scores through learned ensemble weights (Supplementary Note~\ref*{si:note:dms_auto_architecture}).

Across the 217 tested assays, DA-Steer achieved the highest median Spearman $\rho$ among the agents (0.551), with DA-Auto following closely behind (0.531; Figure~\ref{fig:predictive}d).
When all 106 ProteinGym references are scored on the same held-out variants, DA-Steer ranks 2nd among the 113 methods (including ProteinGym baselines and agent references), behind only the supervised Kermut predictor, while DA-Auto ranks 3rd overall~(Supplementary Table~\ref*{si:tab:dms_results}).
DA-Steer also outperforms all baseline agents on this dataset (Figure~\ref{fig:predictive}d).
Coding agent loop (Opus) achieves median $\rho=0.530$ and ranks 5th, while Coding agent loop (Sonnet) achieves $\rho=0.505$ and ranks 9th; AutoDiscovery achieves $\rho=0.313$ and ranks 92nd, while AI-Scientist-v2 and the single-session Opus~4.6 Coding agent@1 rank 112th and 113th, respectively, with negative median Spearman correlations ($-0.056$ and $-0.096$).
Further, DA-Steer's prediction similarities argue against simple reproduction of any single existing variant-effect predictor (Supplementary Note~\ref*{si:note:dms_predictor_similarity}).
DA-Steer was significantly higher than DA-Auto ($p=6.2\times10^{-4}$) and both Coding agent loop (Opus) ($p=0.0059$) and Coding agent loop (Sonnet) ($p=0.0082$), whereas DA-Auto and the two coding-agent-loop variants were not detectably different from one another (all $p\geq0.21$, Supplementary Table~\ref*{si:tab:dms_paired}).
DA-Steer outperforms DA-Auto on 134 of 217 assays (61.8\%), with a mean difference of $\Delta\rho=+0.038$, consistent with the potential benefit of learning interactions among complementary mutation features directly from DMS labels.
DA-Auto remains more competitive on activity assays and proteins with deep alignments, where its pretrained sequence components may benefit from stronger evolutionary signal.
This experiment demonstrates the power of \name to support large-scale predictive modeling development, especially in the scientist-driven case.

\subsection{Audit agents enable more rigorous AI-driven research}
\label{sec:results:audit}

Audit agents, enabled by DiscoveryAgent’s checkpointing and provenance structure, played an integral role in scientists’ ability to steer \name.
We implemented an interface used during the course of the project that supported our hypothesis steering and audit agent functionality (Figure \ref{fig:audit}a, Supplementary Note~\ref*{si:note:audit_ui}).
Through this interface and in post hoc experiments, scientists used audit agents in a variety of ways, including clarifying experimental details, pinpointing exact implementations in the codebase, elucidating project strategy, and reproducing results (Figure \ref{fig:audit}b).
Some observations driven by these interactions are included in "Takeaways" boxes in later sections.
We now illustrate the utility of audit agents through two quantitative experiments as well as an instance of how we used audit agents to prevent test set leakage and to detect disallowed internet downloads.

\begin{figure}[!tp]
    \centering
    \preparefigurecaption{\textbf{\name's provenance structure enables effective auditing.}
    \textbf{a)} Screenshot of the \name hypothesis steering interface. Shown are three candidate hypotheses generated in relation to the case study in Section~\ref{sec:results:perturb}, with Hypothesis 3 selected. At the bottom is an example audit agent chat alongside figures generated by \name.
    \textbf{b)} Example chats with the audit agent; a scientist may ask for analysis details, precise citations to implementations, reasoning on certain decisions, or even to reproduce findings.
    \textbf{c)} Investigation design for studying the ability of agents to identify corrupted claims in traces of different agents. Audit agents are used to audit each of the traces.
    \textbf{d)} Results of agentic corrupted claim identification. Individual points are F1 scores on individual trials; error bars are 95\% $t$-confidence intervals of the mean.
    \textbf{e)} Experiment design for studying the benefit of the audit agent's ability to assist scientists in identifying errors; a research report containing errors was given to two cohorts: code-only (only the report and code generated by the agent) and audit group (given an audit agent attached to \name's research trace).
    \textbf{f)} Total reported findings and unique inserted errors recovered per participant ($n=4$ per cohort). Points represent participants; boxes show the median and interquartile range, with whiskers extending to the most extreme observations within 1.5 times the interquartile range.
    \textbf{g)} Investigation of an earlier round of DA-Auto with the audit agent revealed that \name had downloaded testing data, despite being told to avoid this.
    \textbf{h)} We restored the end-of-Round-5 checkpoint and resumed execution at Round 6, implementing sandbox hardening measures including explicit URL blocking and strong system file isolation. We also performed post hoc validation, using the audit agent, that the final DA-Auto model did not contain leakage.
    }{fig:audit}%
    \includegraphics[width=\linewidth,height=\figureimageheight,keepaspectratio]{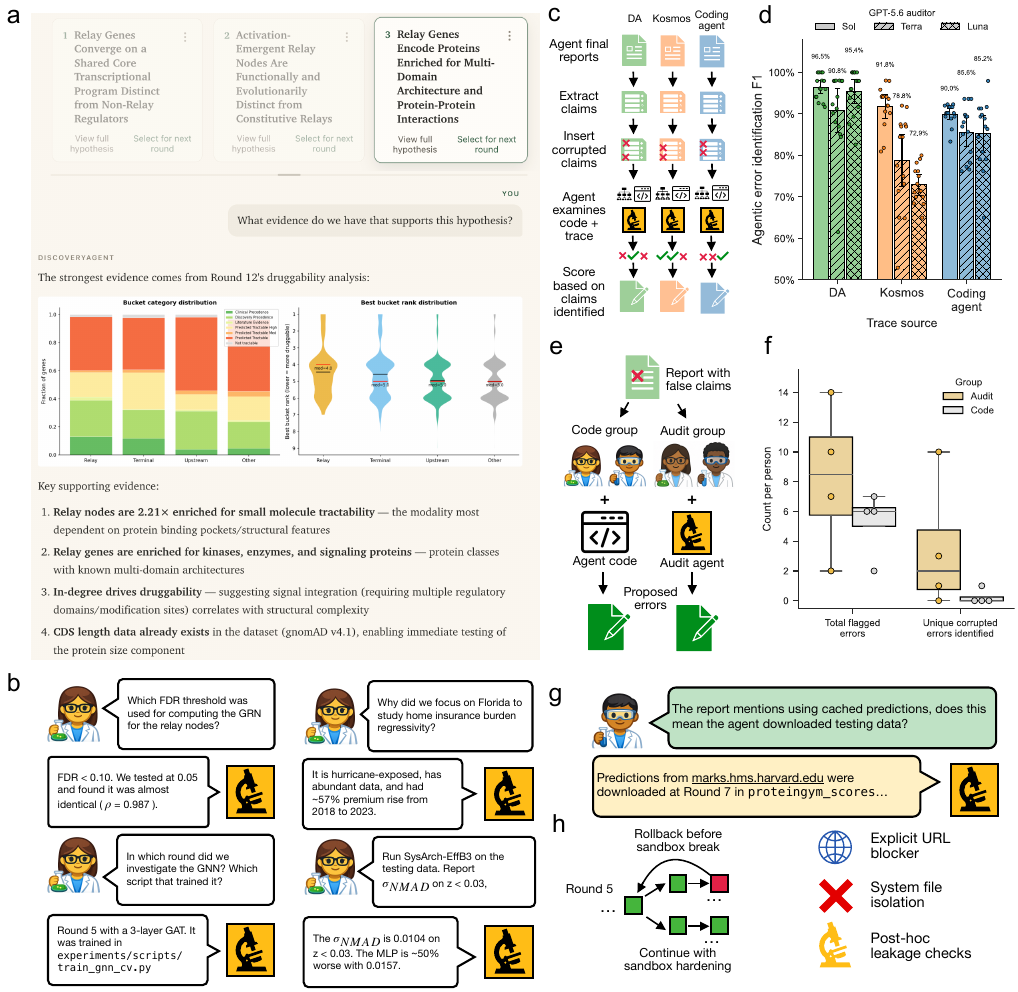}%
    \par\nointerlineskip\usebox{\figurecaptionbox}
\end{figure}

\xhdr{Audit agents aid in identification of errors}
We performed an analysis to understand the ability of audit agents to detect errors in research claims.
Using traces from the StarEmbed dataset, we extracted 50 claims from each of \name and two other agents (Kosmos and the coding-agent loop baseline, or coding agent for brevity), as reported by these individual agents at the end of the research (Figure~\ref{fig:audit}c).
This experiment tests how effectively a fixed auditor configuration, with access to \name's provenance against that of other agents, can retrieve and understand the research trace.
We then corrupted 25 of these claims to artificially inflate results, sampling 5 different sets of claims to corrupt.
The audit agent for \name was then asked to identify the incorrect claims and provide evidence against the claim from the trace; we also asked a coding agent to do the same given the traces of Kosmos and the coding agent.
For the primary comparison, we held the number of inserted corruptions and the auditor harness fixed across systems.
We compare three levels of base models used for the auditors: GPT-5.6-Sol, GPT-5.6-Terra, and GPT-5.6-Luna (Supplementary Note~\ref*{si:note:audit_claim_corruption}).
Kosmos exposed 50 trajectory-specific capsules containing executed notebooks, memory transcripts, and generated artifacts, whereas Coding agent exposed an unstructured cumulative working directory containing chronological interaction logs, analysis code, data, model outputs, and figures.
These structures were extracted, unedited, from each of the agents' available research outputs (Supplementary Note~\ref*{si:note:audit_trace_provenance}).

With GPT-5.6-Sol as the auditor, \name audits achieved near-perfect F1 (0.965) in identifying corrupted claims, with Terra and Luna auditors following closely behind (Figure~\ref{fig:audit}d; Supplementary Table~\ref*{si:tab:audit_claim_f1}).
Coding agents auditing Kosmos and Coding agent traces were lower, with larger drops in audit accuracy for Terra and Luna models (Figure~\ref{fig:audit}d, Supplementary Note~\ref*{si:note:audit_claim_corruption}).
These results hold when controlling for difficulty and type of the corrupted claim (Supplementary Note~\ref*{si:note:audit_corruption_difficulty}, Supplementary Tables~\ref*{si:tab:audit_difficulty_by_system} and~\ref*{si:tab:audit_recall_by_difficulty}).
The errors missed by GPT-5.6-Sol auditors further show that trace availability alone is insufficient, as auditability depends on whether the trace makes relevant evidence easy to locate and disambiguate.
In the Kosmos trace, only 3 of 6 auditors identified that the report had inflated i-band coverage from 44.3\% to 58.7\%, because the relevant dataset summary only pointed to another output containing the true value.
Similarly, only 5 of 9 auditors detected that the report had reduced the confusion rate between two rare classes from 43\% to 34\%, because the correct value appeared hidden inside notebook output rather than in the main results table.
In the coding agent example, all three auditors missed that the report had removed a warning about reusing the validation set for both model selection and final evaluation.
The evidence remained in the trace, but it was scattered across separate experiments rather than linked in a single record.
In another case, the audit agents for the coding agent example recovered inflated rare-class F1 values in only 7 of 9 audits because multiple summaries interleaved metrics from different model versions.
These failure cases illustrate how poorly auditable information structures in alternative agentic systems can cause auditors to miss misrepresentations, such as inflated performance metrics or misrepresented data-corruption warnings.
By organizing evidence into structured checkpoints with accessible intermediate memory synthesis, \name makes supporting evidence easier for a fixed auditor to retrieve and connect to reported claims, resulting in higher corrupted-claim identification performance.
As a negative control, we provided the corrupted reports without research traces; mean F1 for GPT-5.6-Sol decreased to 0.268, 0.135, and 0.315 for \name, Kosmos, and Coding agent, respectively (Supplementary Figure~\ref*{si:fig:audit_report_only_control} and Supplementary Table~\ref*{si:tab:audit_claim_f1}).
Across GPT-5.6-Sol, Terra, and Luna, audits of \name traces retained the highest corrupted-claim identification F1, while report-only audits remained less accurate for all three trace sources.
Importantly, these results are a measure of the ability of audit agents to detect errors in research traces rather than an absolute measure of scientific correctness.

\xhdr{Audit agents assist human scientists in error identification} We next study the ability of audit agents to assist human scientists in error identification.
Using an earlier checkpoint of the DA-Auto run on the ProteinGym dataset, we generate a research paper describing the trace.
We then choose intentional corruptions to insert into the paper that misrepresent results in the codebase and/or agent logs (Methods Sec.~\ref{sec:methods:audit_evaluation}).
We then ask two cohorts of human reviewers to review the paper and identify errors (Supplementary Note~\ref*{si:note:audit_participant_protocol}) within a one-hour limit (Figure~\ref{fig:audit}e).
One cohort is given access to only the paper and the codebase (``Code'' group) and the other cohort is given the audit agents through our audit UI (Methods Sec.~\ref{sec:methods:provenance}, Supplementary Note~\ref*{si:note:audit_ui}) along with the paper, code, and checkpoints (``Audit'' group).
We then measure the total number of errors identified by each cohort along with the proposed errors that match the original inserted errors.

The Audit group reports more findings per participant than the Code group (8.25 versus 5.25) and recovers more unique inserted errors (3.50 versus 0.25; Figure~\ref{fig:audit}f; Supplementary Table~\ref*{si:tab:human_audit_counts}).
Across submitted findings, 14 of 33 in the Audit group match at least one inserted error, compared with 1 of 21 in the Code group (42.4\% versus 4.8\%).
In one audit session, a participant challenged a flagged statistical inconsistency by pointing to supporting figure evidence, prompting the audit agent to distinguish the quantities measured and withdraw its numerical-error claim. 
In another, a participant requested a methodological audit that identified a discrepancy between the paper's stated numerical precision and the implementation, supported by both code and research records, which the participant included in their submitted findings.
In a third example, an audit agent traced DA-Auto's use of cached variant-effect predictions to a download of held-out ProteinGym test data and identified the round in which it occurred (Figure~\ref{fig:audit}g).
We restored the end-of-Round-5 checkpoint, resumed at Round 6 with stronger sandbox controls, and used post hoc audits to verify that no further test-data access occurred in later DA-Auto rounds or DA-Steer (Figure~\ref{fig:audit}h; Supplementary Note~\ref*{si:note:leakage_audit}).
These results, along with the results presented through collaborative efforts by scientists in the following sections, show how \name's design can enable more effective command and understanding of agentic research.

\subsection{Extragalactic astronomy: Photometric redshift prediction}
\label{sec:results:astro}

Redshift ($z$), the shift of a galaxy's light toward longer wavelengths, is used by astronomers to estimate distance from Earth.
Obtaining a galaxy's redshift transforms two-dimensional imaging surveys into three-dimensional maps of the universe.
Spectroscopy yields precise redshifts but requires costly telescope time and specialized instrumentation, so it cannot keep pace with wide-field surveys that image hundreds of millions to billions of sources.
Alternatively, photometric-redshift estimation (photo-$z$) infers redshift from scalable photometry---total brightness in broadband filters ($g$, $r$, $z$) and colors---and images retaining pixel-level information on size, surface brightness, and morphology.
Traditionally, photo-$z$ methods used catalog-level measurements of fluxes, magnitudes, and colors, which are physically related to redshift~\cite{salvato2019many}, but recent machine learning approaches have attempted to estimate photo-$z$ directly from galaxy images~\cite{newman2022photometric, treyer2024cnn}, which leverage a galaxy's morphology for additional expressivity.
Photo-$z$ is challenging because different combinations of galaxy type and distance can produce similar broadband colors, faint or poorly resolved images provide limited morphological information, and nearby (low-$z$) galaxies are rare relative to the much larger background population~\cite{newman2022photometric}.
The low-$z$ regime is particularly important for identifying nearby dwarf galaxies, whose abundances and properties provide sensitive tests of dark matter and stellar feedback but which can resemble more distant galaxies without reliable distance estimates.

We ask \name, steered by a senior astrophysicist, to develop a multimodal estimator that identifies low-redshift galaxies and predicts continuous redshift from catalog photometry and galaxy image cutouts (Figure~\ref{fig:astro}a).
We provide spectroscopic labels from DESI Data Release 1 \cite{DESICollaboration2025DataInstrument} and enrich the research sample for dwarf and nearby systems selected using the criteria of Darragh--Ford et al.\ \cite{darragh2023target}, alongside a broad comparison population.
Many cosmological applications prioritize characterizing ensemble redshift distributions~\cite{newman2022photometric}, whereas our application requires per-galaxy estimates that balance contamination by distant galaxies against completeness in recovering nearby ones. 
Thus, we formulate the problem first as a classification problem where we classify low-$z$ samples (those below a cutoff of 0.03) from high-$z$ samples, and second as a regression problem where we attempt to predict the exact $z$ of the galaxy.
Methods Sec.~\ref{sec:methods:case_studies} and Supplementary Note~\ref*{si:note:dwarf_dataset} describe the task, dataset construction, and evaluation splits.

\begin{casestudytakeaways}{Takeaways from investigation by astronomer + \name}
Driven by an astronomer, \name explored 190 model configurations---spanning model architectures, population resampling, training strategies, etc.---to ultimately develop PhotoFusion, which combines galaxy photometry and images.
The best model achieved 0.971 AUROC for low-redshift classification and reduced photometric-redshift scatter for low-redshift galaxies by 18.6\% relative to the image-only CNN and 39.3\% relative to the photometry-only MLP.
The evaluated joint model showed particularly large error reductions for compact, low-mass galaxies relative to the single-modality baselines.
This collaborative effort developed a joint image--photometry ML predictor, building on established approaches to photometric-redshift estimation, and improved performance relative to baselines.
\end{casestudytakeaways}

\xhdr{Scientist-steered model development}
\name was driven by an astronomer with machine learning experience across 16 rounds (Figure~\ref{fig:astro}b, Supplementary Note~\ref*{si:note:dwarf_trace}).
The agent performed standard preprocessing, exploratory data analysis, and training pipeline preparation in the first three rounds.
Its training scaled over the next few rounds as it attempted advanced ML approaches such as magnitude specialists, domain-specific self-supervision, partially frozen backbones, and residual models, which mostly yielded null results, while exploring trade-offs between the classification and regression formulation of the problem.
In Rounds 6--8, \name and the astronomer investigated the surprising power of a multilayer perceptron (MLP) trained on only five photometry features. In the Round 6 comparison, its validation area under the receiver operating characteristic curve (AUROC) reached $0.888$ and remained stable, whereas early convolutional neural network (CNN) variants that learned image-based residual corrections to the photometric prior peaked earlier at $0.868$--$0.876$ and then declined (Figure~\ref{fig:astro}c).
We use this MLP as the photometry-only baseline throughout the external evaluation below (Supplementary Note~\ref*{si:note:dwarf_single_modality_baselines}).
The next phase involved careful exploration of model architectures, and in Round 9, \name introduced the PhotoFusion architecture, which fused photometry features with a CNN trained on galaxy images (Figure~\ref{fig:astro}d).
These hybrid networks resemble families of models for photometric-redshift estimation that combine image and catalog-photometry branches~\cite{menou2019morpho,henghes2022deep,zhou2022extracting,zhou2022photometric}.
Subsequent rounds tested boundary-aware contrastive learning, checkpoint selection, augmentation, and architecture scaling.
Round 12 investigated a probabilistic model that predicted distributions over the target redshift, and the agent examined where spectroscopic redshifts fell within its predicted distributions (Figure~\ref{fig:astro}e). Probabilistic modeling of photo-$z$ estimation has strong precedent and motivation in the field~\cite{zhou2022photometric}.
Many of the explored features alone provided marginal gains in full-sample performance, but their components supplied the ultimate recipes for the final model.
In Rounds 14 and 15, \name performed wide architecture and hyperparameter sweeps to converge on a final architecture (Figure~\ref{fig:astro}f), and during the final round, the scientist instructed \name to pull together the best discovered training recipes into one group of models.
Over the course of 16 rounds, \name trained 190 model configurations, testing exhaustive components from training data mixtures to architectures and other hyperparameters.

\xhdr{External validation and reproduction} We retrained and evaluated the selected model recipes on a separately constructed sample from the same survey sources (Supplementary Note~\ref*{si:note:dwarf_dataset}; Supplementary Table~\ref*{si:tab:dwarf_dataset_uses}). 
For this sample, we changed the low-redshift classification target from the campaign's $z<0.03$ to $z<0.05$.
The final round consolidated standardized predictions for 125 models from the 190 model configurations trained across the research run, using internal validation for model selection.
Here, we report two of the best-performing models, as chosen from the original development split.
The first model is PF-EffB3, a variant of the PhotoFusion architecture with an ImageNet-pretrained~\cite{deng2009imagenet} EfficientNet CNN backbone \cite{tan2019efficientnet}.
The second model is V3-Binned, a more complex model architecture that used probabilistic heads, MixUp augmentations~\cite{zhang2018mixup}, stochastic depth training~\cite{huang2016deep}, and exponential moving average to prevent overfitting (Supplementary Table~\ref*{si:tab:dwarf_training}).
Both models are of the PhotoFusion lineage, combining catalog photometry and image data to make predictions (Supplementary Note~\ref*{si:note:dwarf_models}).
We retrained both models on the final training split, selected checkpoints on validation, and scored the test split (Supplementary Note~\ref*{si:note:dwarf_external_eval}).
For baselines, we test two single-modality models, including an image-only CNN and the photometry-only MLP developed by \name in Rounds 6--8 (Supplementary Note~\ref*{si:note:dwarf_single_modality_baselines}, Supplementary Note~\ref*{si:note:dwarf_external_eval}).
We also use a previous state-of-the-art low-$z$ classification model from Darragh--Ford et al.~\cite{darragh2023target}.

\begin{figure}[!tp]
    \centering
    \preparefigurecaption{\textbf{\name builds a multimodal photometric-redshift predictor.}
\textbf{a)} Photometric-redshift prediction task.
\textbf{b)} Sixteen research rounds (0--15); highlighted rounds in green.
\textbf{c)} Round 6 validation AUROC for $z<0.03$: photometry-only multilayer perceptron (MLP) versus image-based residual models.
\textbf{d)} Round 9 PhotoFusion comparison by purity at 95\% completeness (P@95C).
\textbf{e)} Round 12 probability integral transform (PIT) distributions for V3-Binned, split at spectroscopic $z=0.03$. PIT is the predicted probability below the spectroscopic redshift. Dashed line denotes uniformity.
\textbf{f)} Round 14 P@95C for 12 architecture variants, evaluated on all of the internal validation set.
\textbf{g)} Low-$z$ ($z<0.05$) classification AUROC (higher is better) and redshift scatter for low-$z$ galaxies and dwarfs (normalized median absolute deviation, $\sigma_{\mathrm{NMAD}}$; lower is better).
\textbf{h)} Mean normalized absolute error for low-$z$ galaxies by stellar mass (dwarfs only) and half-light radius. Gray histograms show counts; bins with fewer than 30 galaxies are omitted.
\textbf{i)} Color--magnitude maps of $\sigma_{\mathrm{NMAD}}$ for joint, image-only, and photometry-only models at $z<0.05$, $0.05\leq z<0.20$, and $z\geq0.20$, with a shared color scale. Rightmost panels show log-scaled counts; cells with fewer than eight galaxies are masked.
\textbf{j)} Photometry-only versus image-only normalized absolute errors, colored by mean PF-EffB3 joint-model error. Highlighted dwarfs illustrate image-only failures (magenta squares), photometry-only failures (yellow circles), and failures of both modalities rescued by the joint model (green diamonds).
    }{fig:astro}%
    \includegraphics[width=\linewidth,height=\figureimageheight,keepaspectratio]{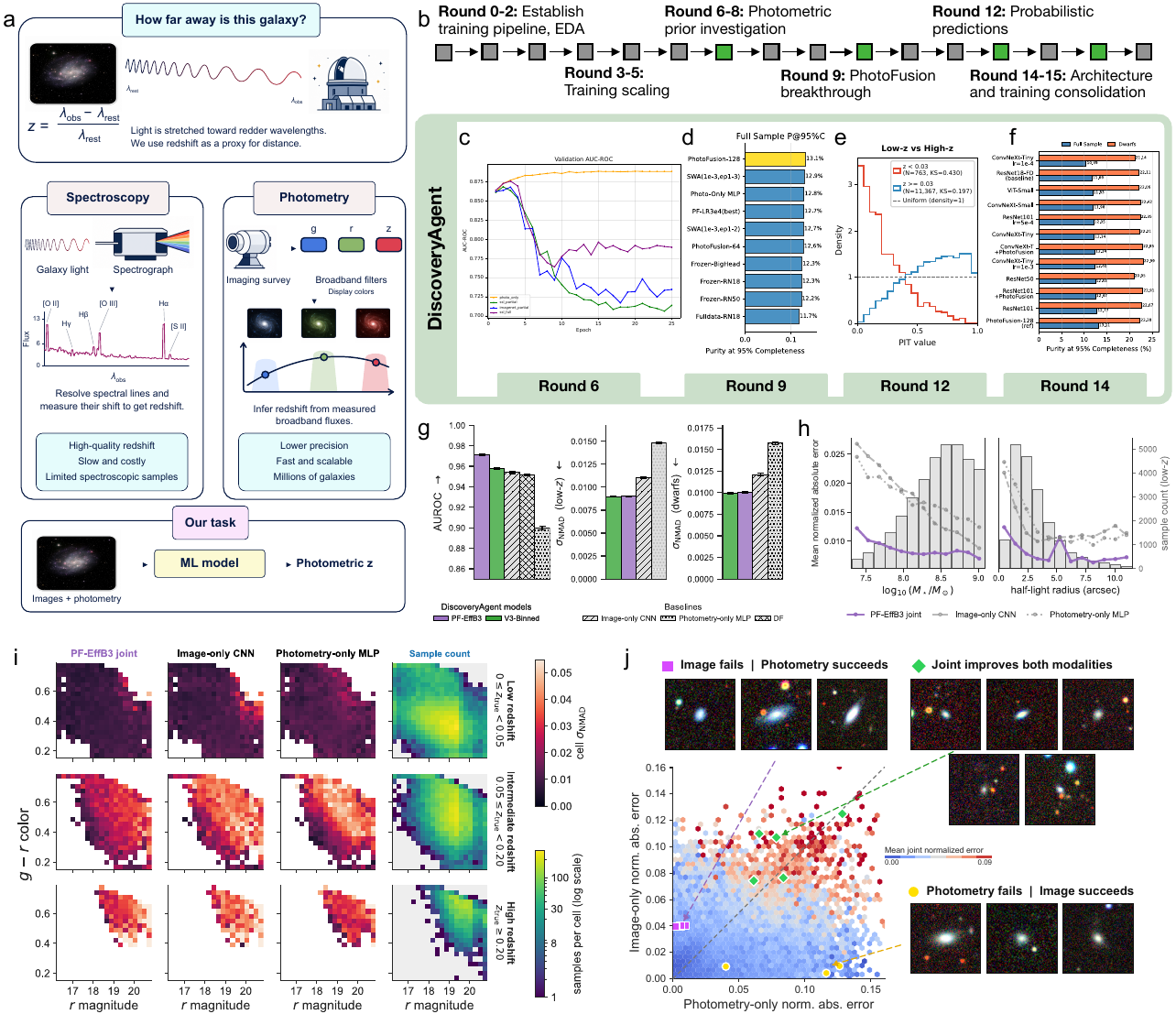}%
    \par\nointerlineskip\usebox{\figurecaptionbox}
\end{figure}

For classification, AUROC measures how reliably a model ranks true low-$z$ galaxies above more distant ones.
For regression, $\sigma_{\mathrm{NMAD}}$ measures the typical scatter between predicted and spectroscopic redshift while ignoring rare catastrophic misses; lower values indicate more precise distance estimates.
On the external held-out split, PF-EffB3 achieved the highest low-$z$ AUROC ($0.971$; $88.4\%$ purity at $95\%$ completeness), while V3-Binned reached $0.958$, compared with $0.954$ for the image-only CNN, $0.951$ for the Darragh--Ford classifier, and $0.900$ for the photometry-only MLP (Figure~\ref{fig:astro}g; Supplementary Note~\ref*{si:note:dwarf_dataset}; Supplementary Table~\ref*{si:tab:dwarf_external_results}).
The AUROC ranking is unchanged after excluding test galaxies shared with the original development set (Supplementary Note~\ref*{si:note:dwarf_overlap}; Supplementary Figure~\ref*{si:fig:dwarf_overlap_sensitivity}).
For low-$z$ galaxies, PF-EffB3 and V3-Binned achieved $\sigma_{\mathrm{NMAD}}$ values of $0.00900$ and $0.00898$, respectively, $18.6$--$18.8\%$ below the image-only CNN ($0.01106$) and $39.3$--$39.4\%$ below the photometry-only MLP ($0.01482$). For dwarf galaxies, their $\sigma_{\mathrm{NMAD}}$ values were $17.4$--$18.3\%$ below the image-only CNN ($0.01211$) and $36.5$--$37.1\%$ below the photometry-only MLP.

We further analyzed the performance of the PF-EffB3 model across underlying properties of the galaxies being predicted.
PF-EffB3 has the lowest low-$z$ mean normalized absolute error across nearly every stellar-mass and half-light-radius bin.
Half-light radius is a galaxy's apparent size on the sky, where galaxies of about one arcsecond (arcsec) are barely resolved in the images.
Photo-$z$ estimates for lower-mass galaxies in our sample show the largest improvement over the single-modality baselines (Figure~\ref{fig:astro}h; Supplementary Table~\ref*{si:tab:dwarf_mae_strata}).
Error also falls sharply with angular size. PF-EffB3 decreases from approximately $0.012$ below $1$ arcsec to mostly $0.006$--$0.007$ above $3$ arcsec, while the baselines remain near $0.009$--$0.013$ over most resolved bins (Figure \ref{fig:astro}h).
The largest relative error reductions compared with these single-modality baselines occur for compact, low-mass low-$z$ galaxies.

\xhdr{Complementary errors support multimodal modeling} We now turn to examining the relative benefits of the multimodal models developed by \name, focusing on PF-EffB3 (``joint model'' for brevity), versus the single-modality baselines.
The joint model, which combines photometry with images that retain galaxy shape and spatial structure, maintains lower error than either single-modality baseline across most populated color--magnitude bins below $z=0.20$ (Figure~\ref{fig:astro}i; Supplementary Table~\ref*{si:tab:dwarf_cmd_summary}).
The image-only and photometry-only models make complementary redshift errors, including image-only errors on visibly resolved galaxies (Figure~\ref{fig:astro}j; Supplementary Figures~\ref*{si:fig:dwarf_joint_kde}, \ref*{si:fig:dwarf_galaxy_predictions}).

We then seek to describe these error patterns as functions of galaxy properties in order to study the contribution of the joint model. 
Across the full held-out test set, the joint prediction was closer to the more accurate single-modality prediction for 71.7\% of galaxies (Supplementary Note~\ref*{si:note:dwarf_modality_following}). 
Its mean normalized absolute error was 8.4\% higher than an oracle that used the true redshift to select the better single-modality prediction for each galaxy. These results suggest that combining images and photometry helps compensate for the limitations of either input alone.

For comparisons of galaxy properties, we define high error as normalized absolute error above each model's test-set median and low error as error at or below that median (Supplementary Table~\ref*{si:tab:dwarf_error_groups}). Galaxies with high photometry-only error but low image-only and joint-model errors were redder and fainter than galaxies with low error from all three models (median $g-r=0.541$ versus $0.422$; $r=19.09$ versus $18.50$).
The yellow-circle examples in Figure~\ref{fig:astro}j illustrate this pattern, with photometry-only predictions overestimating their redshifts and image-only and joint predictions falling closer to the spectroscopic values.
The image-only model underperformed the photometry-only and joint models for very nearby galaxies ($z<0.03$). 
Across all galaxies in this redshift range, its median normalized absolute error was $0.0176$, compared with $0.0112$ for photometry alone and $0.0062$ for the joint model.
The middle and right magenta-square examples in Figure~\ref{fig:astro}j show that image-only overestimates also occur for visibly extended galaxies.

Among galaxies with high error from both single-modality models, those with low joint error were more often nearby ($29.6\%$ versus $10.4\%$ at $z<0.05$) and dwarfs ($30.4\%$ versus $15.1\%$) than those with high joint error (Supplementary Table~\ref*{si:tab:dwarf_error_groups}).
The green-diamond examples in Figure~\ref{fig:astro}j show nearby dwarfs that both single-modality models place at much higher redshifts, while the joint predictions remain close to their spectroscopic redshifts.
The joint model nevertheless had higher error than both single-modality baselines for $17.4\%$ of test galaxies; the complete per-galaxy comparisons, residual statistics, and eight-group decomposition are reported in Supplementary Tables~\ref*{si:tab:dwarf_joint_pairwise_wins}, \ref*{si:tab:dwarf_residual_structure}, and~\ref*{si:tab:dwarf_error_groups}.
These examples are consistent with images and photometry providing complementary information for redshift prediction.
Together with the external performance gains, these data support the decision arrived at by the astronomer and \name to develop the PhotoFusion architecture to address the limitations of photometry and image data alone.

\begin{steeringobservations}{Astronomer's perspective on working with \name}
Across the case study, the collaboration evolved from broad exploration to targeted model refinement.
Early rounds required the astronomer to repeatedly reframe the scientific objective, especially whether to prioritize low-redshift classification or redshift regression, while \name explored dataset configurations, architectures, objectives, augmentation schemes, and optimization strategies at a rate much greater than a single scientist's capacity.
Once PhotoFusion emerged, the astronomer's role shifted toward interpreting positive and negative results, selecting follow-up hypotheses, and judging when the search had converged.
This process challenged prior assumptions and clarified when photometry and morphology provide complementary information, why gains in classification do not necessarily yield more accurate redshift estimates, and where model performance plateaus.
Autonomy between rounds created a productive framework for human engagement, reducing approval fatigue while enabling broad experimentation.
Between rounds, the scientist could inspect evidence, interpret failures, consult collaborators, and refine the next hypothesis, using audit agents to interrogate underlying analyses when needed.
The final models therefore emerged through co-learning where \name broadened the empirical search, while the astronomer reframed the problem, interpreted the evidence, focused the investigation, and determined when to conclude.
\end{steeringobservations}

\subsection{Economics: Assessing insurance costs as a component of housing affordability}
\label{sec:results:econ}

We next evaluate \name on an open-ended discovery task in empirical economics, where the goal is not to optimize a prediction but to surface descriptive insights.
This analysis illustrates the kinds of descriptive analyses that often serve as components of empirical economic research.
We tasked \name to explore the American Community Survey (ACS) and identify novel, interesting phenomena (Methods Sec.~\ref{sec:methods:case_studies}).
To acquire data, we developed and provided \name with a Census data agent with search-and-download access to a broad collection of public-use U.S. Census and federal survey data (Figure~\ref{fig:economics}a, Supplementary Note~\ref*{si:note:census_data}, Supplementary Table~\ref*{si:tab:acs_table_series}).\footnote{These descriptive estimates rely on the American Community Survey and are therefore subject to the usual limitations of large-scale survey data, including sampling error, nonresponse, measurement error, and potential selection and reporting biases.}
Over multiple rounds of hypothesis steering, the economist and \name explored the affordability of housing insurance in markets acutely stressed by rising climate risk, converging on Florida as a case study.

\begin{casestudytakeaways}{Takeaways from investigation by economist + \name}
Steered by an economist, \name identified a sharp rise in Florida homeowners-insurance costs between 2018 and 2023.
This burden fell increasingly on lower-income homeowners: between 2018 and 2023, the percentage-point increase in insurance's share of income was $\approx7.1$ times as large for the lowest income quartile as for the highest.
These findings are consistent with prior national evidence and suggest that rising insurance costs are making homeownership less affordable in Florida, supporting the treatment of insurance as a core variable alongside home prices and mortgage payments when assessing affordability.
\end{casestudytakeaways}

\xhdr{From an open data universe to a focused hypothesis} Starting from this broad data collection (Fig.~\ref{fig:economics}a), the economist and \name progressively narrowed to focused questions on home insurance affordability (Fig.~\ref{fig:economics}b).
In Round 1, using 2023 ACS 1-year PUMS housing microdata, \name compared the insurance-to-income ratio across three hurricane-exposed states (Louisiana, Florida, Texas) and three states with less exposure (Minnesota, Ohio, California).
This analysis found that the burden was both higher and more dispersed in hurricane-prone states, with Louisiana and Florida owners spending roughly twice the income share on insurance as owners in Minnesota or Ohio (Fig.~\ref{fig:economics}c, Supplementary Table~\ref*{si:tab:econ_round1_states}).
Narrowing to Florida in Round 6, \name used the 2018 and 2023 ACS 1-year PUMS housing samples to document that the market had shifted sharply within five years.
The weighted-median annual premium rose from \$1{,}400 in 2018 to \$2{,}200 in 2023 (+57\%), with the entire premium distribution translating rightward (Fig.~\ref{fig:economics}d).

\xhdr{Independent reproduction confirms a regressive, worsening burden} Using the audit agents, we reanalyzed and extended these ACS PUMS findings across additional years and geographies. 
First, the Florida surge is stronger than the national trend, with an 83\% rise from 2013 to 2023, 
outpacing the national median (\$800 to \$1{,}300) and median household income growth over the same period (+52\%) (Fig.~\ref{fig:economics}e, Supplementary Table~\ref*{si:tab:econ_trends}), consistent with prior evidence that premiums have outpaced income nationally and grown fastest in high-risk markets~\cite{keys2024property, zillow2025insurance}. 
Premiums are primarily priced to the property and its risk rather than the income of the homeowner, so the same cost weighs far more heavily on smaller incomes.
In 2023, the lowest income quartile spent 6.6\% of income on insurance compared with 1.2\% for the highest---up from 4.9\% and 1.0\% in 2018, a 1.7-percentage-point increase for the lowest-income households but only 0.2 for the highest (Fig.~\ref{fig:economics}f, Supplementary Table~\ref*{si:tab:econ_burden}).
As with federal flood insurance \cite{kousky2018financing}, and consistent with prior ACS-based evidence that the heaviest housing-cost burdens fall on lower-income owners~\cite{jchs2024acs, zillow2025insurance}, these results show that homeowners insurance burden appears to be regressive with respect to income. Because some low-income owners (e.g., retirees) hold substantial housing wealth, this income-based measure need not imply regressivity in a broader welfare sense.
This finding also complements national estimates from mortgage-escrow data \cite{keys2024property, urban2026property} by covering owner-occupied households reporting positive insurance expenditures regardless of mortgage status, including owners who hold their homes free and clear and are absent from mortgage-based data; ACS tabulations have similarly documented rising insurance-driven costs for mortgage-free owners~\cite{jchs2024acs}.

\begin{figure}[!tp]
    \centering
    \preparefigurecaption{\textbf{\name and economist team elucidate impact of home insurance in hurricane-exposed states}
    \textbf{a)} The agent was given access to a broad collection of public-use U.S. Census and federal survey data: ACS summary tables, confidentiality-protected ACS PUMS microdata, and harmonized IPUMS archives.
    \textbf{b)} \name discovery driven by an economist. Highlighted rounds shown with a green box; other rounds in a grey box.
    \textbf{c)} In Round 1, \name investigated home insurance affordability, finding it disproportionately burdens hurricane-exposed states. Boxplot shows the insurance-to-income ratio across 6 states: 3 hurricane-prone states (Louisiana, Florida, Texas) and 3 states less exposed to hurricanes (Minnesota, Ohio, California).
    \textbf{d)} In Round 6, \name narrowed to Florida and found the weighted-median annual premium rose from \$1{,}400 (2018) to \$2{,}200 (2023), a 57\% increase, with the full distribution shifting rightward.
    \textbf{e)} As a reproduction, median insurance premium in Florida rose from \$1{,}200 (2013) to \$2{,}200 (2023), outpacing the national median (\$800 to \$1{,}300) and median household income growth.
    \textbf{f)} Reproduction confirming that the burden is regressive with respect to income. Insurance as a percentage of income (ITI) by income quartile, with shifts from 2018 to 2023. In 2023, the lowest income quartile paid 6.6\% of income on insurance compared with 1.2\% for the highest, up from 4.9\% and 1.0\% in 2018, respectively.
    \textbf{g)} Cumulative growth as a percentage of 2013 cost over time. Insurance grows 83\%, faster than every other ownership cost, with a large increase in 2022--2023. Dashed lines mark major landfalling hurricanes (Irma, Michael, Ian) for reference; we do not attribute the increase to any single driver.
    \textbf{h)} Capitalizing PUMA-level insurance costs at 7\% moves 41 of 168 PUMAs from raw P/R $\leq 21$ to adjusted P/R $>21$. The connected axes show each PUMA's shift; shaded interval spans P/R 15--21, and orange lines mark PUMAs crossing the stricter threshold of 21. The map locates these transition PUMAs (orange) relative to PUMAs already above the threshold before capitalization (muted orange) and those remaining at or below it after capitalization (blue).
    }{fig:economics}%
    \includegraphics[width=\linewidth,height=\figureimageheight,keepaspectratio]{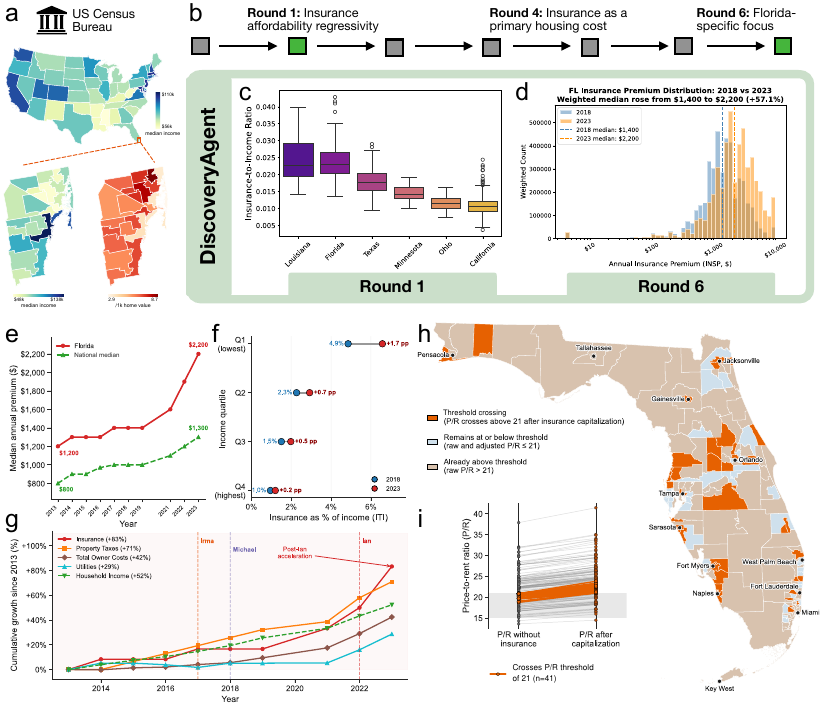}%
    \par\nointerlineskip\usebox{\figurecaptionbox}
\end{figure}

\xhdr{Insurance burden shifts modeled price-to-rent thresholds and coverage decisions} Insurance has become the fastest-growing component of Florida ownership cost, rising 83\% since 2013---ahead of property taxes (+71\%), total owner costs (+42\%), utilities (+29\%), and household income (+52\%)---with a sharp acceleration in 2022--2023, a period that followed major hurricanes (Irma, Michael, Ian) but also saw litigation-driven losses, rising reinsurance costs, and insurer insolvencies; we do not attribute the increase to any single driver (Fig.~\ref{fig:economics}g; Supplementary Table~\ref*{si:tab:econ_trends}).
Given forecasts of more intense and wetter hurricanes, a rising proportion of Category~4--5 storms~\cite{knutson2020tropical, ipcc2021physical}, and increased coastal hurricane frequency over the Florida peninsula~\cite{balaguru2023increased}, this affordability pressure may persist or even worsen. 

The increase is already large enough to change modeled tenure decisions.
We used audit agents, working off of \name's initial exploration, to perform a downstream, targeted analysis on the implication of home insurance on modeled buy vs. rent decisions.
The price-to-rent ratio (P/R) is sometimes used as a heuristic for whether a market is more suitable for buying or renting a home.
Trulia's 2012 consumer-market heuristic favors buying at P/R at or below 15 and renting at P/R at or above 20\cite{trulia2012rentvsbuy}; we use these reference values as descriptive benchmarks. 
Between 15 and 20 is considered a ``grey zone'' where the advantage to buy or rent is less clear for a buyer.\footnote{This cutoff is considered a ``rule-of-thumb'' and is used here as illustrative of large shifts in housing affordability due to capitalizing insurance.}
In this analysis, we report a P/R of 21 as a stricter threshold, where above 21 is considered to be strongly rent-favored.
Capitalizing local insurance costs pushed 41 of 168 Florida Public Use Microdata Areas (PUMAs; $24.4\%$) across our stringent P/R threshold of 21 
($\approx14\%$ at the cutoff of 20; Figure~\ref{fig:economics}h, Supplementary Note~\ref*{si:note:econ_buyrent}, Supplementary Table~\ref*{si:tab:econ_pr_counts}).
Rising insurance costs meaningfully increase the P/R ratio, potentially shifting the implied tenure decision from buying toward renting.
This is especially impactful because simple price-to-rent heuristics typically omit insurance, and user-cost frameworks, while including recurring ownership costs, typically do not model its large spatial variation~\cite{himmelberg2005assessing, poterba1984tax}.
These transition PUMAs span the Panhandle, central Florida, and several coastal metropolitan areas rather than clustering in the most expensive markets, which were already above the threshold before capitalization (Fig.~\ref{fig:economics}h).
A caveat worth noting is that this analysis capitalizes insurance into ownership cost mechanically, but in equilibrium, rising premiums may also change both home price and rental price growth in exposed markets~\cite{keys2024property}, which would change the trajectory of the P/R shift over longer horizons.\footnote{Note that this analysis does not capitalize the insurance cost into rental prices. Because rising homeowners insurance costs can affect both the numerator (home prices) and denominator (rents) of the P/R ratio in equilibrium, their net effect on the ratio is theoretically sign-ambiguous.}
However, this analysis shows that homeowners insurance costs might have a meaningful effect on home affordability in Florida, to the point that a large number of regions become more rent-favorable after capitalizing insurance than without.

High insurance costs may also affect whether current owners choose to purchase insurance at all. 
Because the premium and burden estimates above condition on positive reported insurance expenditure, we separately examined all eligible Florida owner-occupied households. 
In 2023, 17.1\% reported zero insurance expenditure; this rate reached 30.6\% in the lowest income quintile compared with 8.1\% in the highest, and 28.5\% among mortgage-free owners compared with 7.6\% among mortgage holders. 
This rate was relatively constant from 2018 to 2023 (17.3\% in 2018; Supplementary Note~\ref*{si:note:econ_going_bare_trend}), so over this window the ACS does not show owners dropping coverage as premiums rose.
Zero reported expenditure is a proxy rather than definitive evidence of uninsurance because it may also reflect reporting error or an escrowed premium not reported separately. 
Nonetheless, the concentration of zero expenditure among lower-income and mortgage-free owners is consistent with national American Housing Survey evidence that uninsurance is concentrated among these groups~\cite{cfa2024exposed}, and identifies a second margin of vulnerability beyond the buy-versus-rent decision (Supplementary Note~\ref*{si:note:econ_going_bare}, Supplementary Table~\ref*{si:tab:econ_burden}).

These findings describe a compounding dynamic, where insurance burdens are heaviest and have grown most where homeownership is already most fragile---lower-income owners, mortgage-free owners, and hurricane-exposed markets---producing a burden that is regressive, worsening, and large enough to potentially shift homeownership decisions.
Taken together, this suggests that rule-of-thumb affordability and buy-versus-rent comparisons should incorporate local insurance costs for a climate-risk era.
The contribution of this investigation, driven jointly by \name and an economist, is to surface and systematically quantify these interlocking patterns.

\begin{steeringobservations}[unbreakable]{Economist's perspective on working with \name}
As opposed to the relatively autonomous role of \name in the extragalactic astronomy case study, \name played a much more exploratory role in this analysis, yielding more autonomy to the expert in the downstream analyses and formalization of results.
The economist noted that \name tended to make stronger assumptions than would typically be made during a descriptive analysis.
\name veered into attempting to make causal inferences, which required heavy steering from the economist, inserting consistent guidance and corrections to the agent between rounds.
Audit agents played a large role in reproducing and running post hoc analyses, understanding which components of the analysis could yield transferrable results.
While such behaviors were less suited to the desired type of analysis, the tight audit agent and hypothesis steering by the economist helped formulate the research findings, despite the priors of the LLMs used in \name being more fixed towards causal rather than descriptive analyses.
The iterative and feedback-centric nature of \name allowed the system and economist to converge on an interesting empirical finding.
\end{steeringobservations}

\subsection{Functional genomics: Prioritizing autoimmune risk genes from perturbation phenotypes}
\label{sec:results:perturb}

Genome-wide association studies (GWAS) implicate hundreds of loci in autoimmune disease, but association alone identifies neither the causal gene nor how a variant affects cellular biology, complicating variant interpretation.
Perturb-seq~\cite{dixit2016perturb} measures transcriptional responses to in vitro gene knockdown, providing functional readouts and inference of regulatory circuits.
However, Perturb-seq alone does not establish which perturbation effects are relevant to human disease.
GWAS and Perturb-seq are therefore complementary---genetics identifies disease-associated loci, while perturbation helps reveal what those genes do within the cell.
Integrating these modalities can help connect disease loci to candidate genes and cellular programs.
The genome-scale Perturb-seq atlas of Zhu et al.~\cite{zhu2025genome} measures transcriptional responses to gene knockdown in primary CD4$^{+}$ T cells (Figure~\ref{fig:gw_cd4}a), which coordinate adaptive immunity and contribute to autoimmune disease~\cite{lagattuta2024genetic}.
Guided by a computational biologist, we used \name to integrate these perturbation profiles with genetic evidence and test how they relate to autoimmune risk.

\begin{casestudytakeaways}{Takeaways from investigation by computational biologist + \name}
Together with a computational biologist, \name identified CD4$^{+}$ T-cell ``relay genes'' as highly regulated genes that \textit{also} broadly influence other genes.
Relay genes were more likely than other genes to be linked to autoimmune diseases and inherited immune deficiencies, and they were more often annotated as potential targets for small molecule drugs.
Combining gene-perturbation experiments with human genetic evidence prioritized four autoimmune-associated genes whose roles in disease remain poorly understood and suggested how they might affect T-cell function, providing concrete hypotheses for follow-up experiments.
\end{casestudytakeaways}

\xhdr{\name exploration trajectory} The \name research trajectory was shaped by a series of decisive negative results that narrowed the search space toward network topology (Figure \ref{fig:gw_cd4}b).
To begin the analysis, we gave \name access to differential expression statistics from the Zhu et al. atlas \cite{zhu2025genome} (Methods Sec.~\ref{sec:methods:case_studies}).
\name was augmented with a sub-agent specialized in OpenTargets, exposing a large amount of aggregated evidence across diseases and genes~\cite{ochoa2023next, zhang2026virtual} (Supplementary Note~\ref*{si:note:opentargets_agent}).
In Round 4, we also added BioMCP~\cite{biomcp_2026}, which exposes over 40 common biomedical databases.
\name was driven in hypothesis steering mode by a human expert with expertise in human genetics and Perturb-seq analysis.
The agent was asked to characterize the perturbation phenotypes of genes that are both perturbed in the Perturb-seq data and implicated in autoimmunity by GWAS~\cite{farh2015genetic, kanai2018genetic}.

Early exploratory rounds involved testing multiple hypotheses about autoimmune GWAS genes that were rejected.
\name found in Round 2 that putative causal genes did not have higher out-degree than neighboring candidates within the same GWAS locus; in Round 3, however, these genes showed higher out-degree than the genome-wide background of perturbed genes.
In Round 5, \name proposed the relay gene hypothesis, stating that genes serving as both heavily regulated targets (high in-degree) and broad upstream regulators (high out-degree) are disproportionately disease-associated.
This was confirmed in Round 5 with a reported odds ratio of 2.91 for GWAS enrichment among relay genes vs. other genes (Figure~\ref{fig:gw_cd4}c; Supplementary Table~\ref*{si:tab:gwcd4_round5_hitrate}).
In Rounds 6--7, the original analyses detected no additional GWAS predictive value from feedback loops or directional asymmetry beyond relay status (Figure \ref{fig:gw_cd4}d, Supplementary Table~\ref*{si:tab:gwcd4_round7_dai}).
The principle was then strengthened on primary immunodeficiency (PID) genes, which are implicated by rare Mendelian variants (Round 8), unifying disease-association evidence from common and rare genetic variants.
Additionally, selection constraints analysis revealed that the most extreme relay genes are under higher evolutionary constraint (measured by $s_{\text{het}}$~\cite{zeng2024bayesian}) than other categories (Round 9, Figure \ref{fig:gw_cd4}e, Supplementary Table~\ref*{si:tab:gwcd4_round9_shet}).
Cross-cell-type validation in a K562 Perturb-seq atlas \cite{replogle2022mapping} showed that the link between relay status and selective constraint is conserved, whereas autoimmune GWAS enrichment did not replicate (Round 10).
Finally, \name found in Round 12 that relay genes were highly enriched for druggable proteins (Figure \ref{fig:gw_cd4}f, Supplementary Table~\ref*{si:tab:gwcd4_round12_tractability}).

\begin{figure}[!tp]
    \centering
    \preparefigurecaption{
    \textbf{\name and a computational biologist identify relay genes.}
    \textbf{a)} Genome-scale CRISPR interference (CRISPRi) Perturb-seq measures transcriptional responses to gene knockdown in primary CD4$^{+}$ T cells~\cite{zhu2025genome}.
    \textbf{b)} Research trajectory summary; highlighted rounds in green.
    \textbf{c)} Round 5 autoimmune genome-wide association study (GWAS) hit rates by network degree (right). The scatter (left) uses an early relay cutoff (both degrees $\geq50$), which was later updated to upper-quartile cutoffs.
    \textbf{d)} Round 7: regulatory directionality varies with feedback similarly in GWAS and non-GWAS genes.
    \textbf{e)} Round 9: relay genes show greater evolutionary constraint ($s_{\mathrm{het}}$).
    \textbf{f)} Round 12: relay enrichment for small-molecule tractability and clinical precedence.
    \textbf{g--h)} Network roles and examples: relay (STAT3), upstream (VAV1), and terminal (IFNG). In-degree counts regulators; out-degree counts affected genes. Relay genes occupy both upper quartiles, upstream genes only the out-degree upper quartile, and terminal genes only the in-degree upper quartile. Neighborhoods show regulators at left and affected genes at right.
    \textbf{i)} Relay enrichment for autoimmune GWAS, primary-immunodeficiency (PID), and erythrocyte-trait genes in CD4$^{+}$ and K562 cells. Points and whiskers show log odds ratios (OR) and 95\% Woolf confidence intervals; stars indicate uncorrected one-sided Fisher exact tests ($\ast\,p<0.05$, $\ast\ast\,p<0.01$, $\ast\ast\ast\,p<0.001$).
    \textbf{j)} Small-molecule tractability enrichment using an independent gene set~\cite{minikel2024refining}.
    \textbf{k)} Functional enrichment across network roles. Simplex positions show relative membership and point sizes indicate OR; right-hand plots show each role's ten most enriched functions. Colors denote relay (yellow), upstream (green), terminal (blue), and shared relay--terminal (purple).
    \textbf{l)} Filtering autoimmune-associated relay genes prioritizes less-characterized candidates; examples of excluded genes appear at upper right. TUFM and ATXN2L illustrate upstream regulators and downstream perturbation responses (lower right).
    }{fig:gw_cd4}%
    \includegraphics[width=\linewidth,height=\figureimageheight,keepaspectratio]{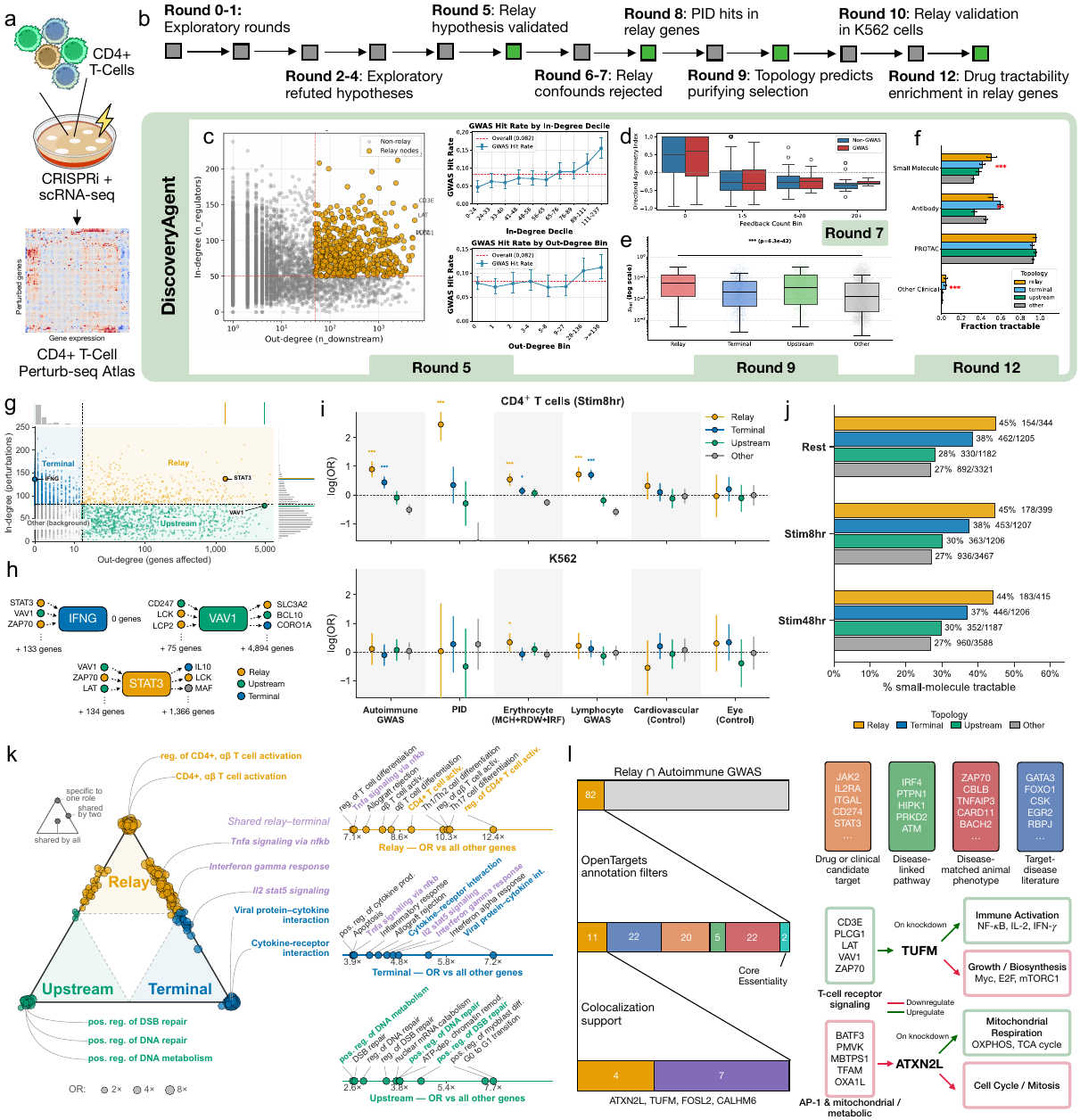}%
    \par\nointerlineskip\usebox{\figurecaptionbox}
\end{figure}

\xhdr{Defining and illustrating relay genes} The central finding from the research trace was that ``relay genes'', those in the upper quartiles of both in-degree and out-degree in the GRN, are robustly enriched for autoimmune disease GWAS hits.
\name made the distinction between relay genes, upstream genes, and terminal genes.
Relay genes are in the upper quartile of both in-degree and out-degree; upstream genes are in the upper quartile of out-degree only; and terminal genes are in the upper quartile of in-degree only (Figure~\ref{fig:gw_cd4}g, Supplementary Note~\ref*{si:note:gwcd4_relay_standardization}).
For illustration, we select three representative examples (Figure~\ref{fig:gw_cd4}h, Supplementary Note~\ref*{si:note:gwcd4_exemplar_nodes}, Supplementary Table~\ref*{si:tab:gwcd4_exemplar_nodes}).
IFNG is terminal (in-degree $136$, out-degree $0$): its transcription integrates many inputs, but secreted IFN-$\gamma$ acts on other cells~\cite{balasubramani2010modular}.
VAV1 is upstream (in-degree $78$, out-degree $4{,}897$): this proximal T-cell-receptor regulator is controlled largely through phosphorylation, producing a broad transcriptional footprint despite fewer transcriptional inputs~\cite{bustelo1992product}.
Finally, STAT3 is a relay gene (in-degree $137$, out-degree $1{,}369$): this cytokine-responsive transcription factor integrates immune signals into broad CD4$^{+}$ T-cell programs \cite{mathur2007stat3}.
STAT3 also has many known roles in autoimmune disease with mechanistic and genetic evidence~\cite{wu2026targeting}.

\xhdr{Reproducing relay gene hypothesis with audit agents} We used audit agents to reproduce the central claims of this research trace.
We first validate that autoimmune GWAS genes are indeed enriched in CD4$^{+}$ T-cell relay genes (OR$=2.44$), and that PID genes are enriched even more strongly (OR$=11.66$); these are computed by fixing reference autoimmune GWAS and PID sets (Figure \ref{fig:gw_cd4}i; Supplementary Note~\ref*{si:note:gwcd4_topology_enrichment}, Supplementary Figure~\ref*{si:fig:gwcd4_all_conditions_forest}, Supplementary Table~\ref*{si:tab:gwcd4_topology_enrichment}).
This enrichment is also robust across multiple autoimmune diseases (Supplementary Figure~\ref*{si:fig:gwcd4_autoimmune_disease_enrichment}) and across the OpenTargets locus-to-gene (L2G) cutoff used to define the autoimmune GWAS gene set (Supplementary Figure~\ref*{si:fig:gwcd4_l2g_sensitivity}).
While K562 relay genes show a nominally significant enrichment for the erythrocyte traits matched to their lineage~\cite{ota2026causal} (one-sided $p=0.033$; not significant after multiple-testing correction), this enrichment is largely specific to the immune context.
Relay genes from three non-immune Perturb-seq atlases (K562, HCT116, and HEK293T) show no enrichment for autoimmune or PID genes, and control cardiovascular and eye-disease GWAS sets are not significantly enriched among relay genes in any tested cell type (Supplementary Note~\ref*{si:note:gwcd4_celltype_specificity}, Supplementary Figure~\ref*{si:fig:gwcd4_celltype_specificity}).

We then turned to reproducing the druggability analysis.
Because the original Round 12 analysis used several definitions of druggability, we fixed a single definition using an independent tractability set from Minikel et al.~\cite{minikel2024refining}.
With this set, relay genes were consistently enriched for small-molecule tractability in all three conditions (Figure~\ref{fig:gw_cd4}j, Supplementary Figure~\ref*{si:fig:gwcd4_druggability_categories}, Supplementary Table~\ref*{si:tab:gwcd4_druggability_annotation}).
However, we found that this phenomenon is attenuated when accounting for the annotation depth of the genes, with only the small-molecule association holding after this adjustment (Supplementary Note~\ref*{si:note:gwcd4_druggability_annotation}).
Audit agents helped to qualify this claim by reproducing this analysis and accounting for annotation as a confounder, which was not considered during the original \name trace.

\xhdr{Relay genes show distinct functional profiles}
Relay, terminal, and upstream genes occupy separable functional compartments in the network (Figure~\ref{fig:gw_cd4}k; Supplementary Table~\ref*{si:tab:gwcd4_role_terms}).
Among 486 terms enriched against the 10,282-gene background, 267 were most strongly enriched in relay genes, 180 in terminal genes, and 39 in upstream genes; their concentration near the simplex corners indicates role-specific biology, with most sharing restricted to the relay--terminal axis.
The strongest terms distinguish these roles.
Relay genes are enriched for CD4$^{+}$ T-cell activation (for example, regulation of CD4$^{+}$, $\alpha\beta$ T-cell activation, OR$=12.4$), terminal genes for cytokine-receptor programs (for example, viral protein--cytokine receptor interaction, OR$=7.2$), and upstream genes for genome maintenance (for example, DNA repair, OR$=3.8$).
Shared relay--terminal terms include TNF$\alpha$--NF-$\kappa$B, IFN-$\gamma$ response, and IL2--STAT5 signaling, consistent with activation programs being integrated by relay genes and executed through terminal outputs, whereas the strongest upstream terms are upstream-specific.
Relay functions are also the most stimulation-dependent, shifting from a poised lymphoid program at Rest to acute T-cell activation at 8\,hr and committed lineage and checkpoint regulation by 48\,hr (Supplementary Figure~\ref*{si:fig:gwcd4_relay_pathway_rank}; Supplementary Note~\ref*{si:note:gwcd4_role_function}).
Together, these results show that considering joint in- and out-degree identifies functionally distinct regulatory layers that neither degree alone reveals.

\xhdr{Mechanistic characterization of poorly studied relay genes} We then used audit agents to help investigate if relay topology can help prioritize autoimmune-associated genes with limited mechanistic evidence (Figure~\ref{fig:gw_cd4}l).
Starting from 82 autoimmune GWAS-linked relay genes, we excluded DepMap common-essential genes (Supplementary Note~\ref*{si:note:gwcd4_essentiality}, Supplementary Figure~\ref*{si:fig:gwcd4_essentiality}) and those with existing drug, pathway, animal-model, or literature evidence, leaving 11 candidates (Figure~\ref{fig:gw_cd4}l, left; Supplementary Note~\ref*{si:note:gwcd4_filtration}).
These filters removed established autoimmune regulators and drug targets---including JAK2, IL2RA, and ZAP70---to focus follow-up on less-characterized candidates (Figure~\ref{fig:gw_cd4}l, upper right).
Colocalization between molecular quantitative trait loci and disease GWAS signals further narrowed the set to TUFM, CALHM6, ATXN2L, and FOSL2, all supported by inflammatory bowel disease signals in immune-relevant cell types (Supplementary Note~\ref*{si:note:gwcd4_coloc}).
Perturb-seq then connected these candidates to upstream regulators and downstream transcriptional programs (Figure~\ref{fig:gw_cd4}l, lower right).
TUFM, an ulcerative-colitis-associated mitochondrial elongation factor, was positively regulated by proximal T-cell-receptor components, including CD3E, PLCG1, and LAT; its knockdown increased NF-$\kappa$B, IL-2/STAT5, and IFN-$\gamma$ response programs while decreasing Myc, E2F, and mTORC1 growth programs.
ATXN2L, a Crohn's-disease-associated RNA-binding and stress-granule-related protein, was negatively regulated by AP-1 and mitochondrial/metabolic factors; its knockdown increased mitochondrial-respiration programs while decreasing cell-cycle and mitotic programs (Supplementary Note~\ref*{si:note:gwcd4_targets}; Supplementary Table~\ref*{si:tab:gwcd4_candidate_programs}).
Although TUFM and ATXN2L are immediate neighbors at 16p11.2, their downstream neighborhoods overlap minimally at Stim8hr (Jaccard $=0.012$) and their perturbation signatures are slightly anticorrelated ($r=-0.11$), supporting distinct functions (Supplementary Note~\ref*{si:note:gwcd4_targets}).
Together, these profiles provide testable functional hypotheses for less-characterized autoimmune-associated genes, although their contribution to disease remains to be established.

\xhdr{\name drives unique hypothesis selection} For comparison, we asked Claude Code with Claude Opus 4.6 to explore the same dataset and generate 100 hypotheses.
Only one proposal combined incoming and outgoing connectivity, predicting that genes with both properties would identify the causal gene among candidates at an autoimmune GWAS locus. 
Our analyses with \name did not support this within-locus prediction. 
Instead, our subsequent analyses established relay genes as a distinct regulatory class enriched for autoimmune associations beyond incoming connectivity alone, with functional profiles consistent with integrating regulatory inputs and coordinating T-cell activation programs. 
This nuance is biologically consequential because relay topology identifies a disease-enriched regulatory class that can guide mechanistic follow-up, but this enrichment does not establish which gene mediates an individual GWAS association.
Thus, iterative investigation with \name refined a broad centrality intuition into a specific account of disease-associated regulatory roles and established the limits of using those roles for causal-gene prioritization.

\begin{steeringobservations}[unbreakable]{Computational biologist's perspective on working with \name}
The scientist used \name as an exploratory collaborator to operationalize broad hypotheses and test them against Perturb-seq and human-genetics data.
\name enabled rapid evaluation of many hypotheses, including negative results that helped determine which directions merited further analysis.
They described the interaction as resembling supervision of an early-stage graduate student: \name required guidance on high-level direction and occasional correction of its assumptions, but it competently executed scoped analyses.
As a complement, audit agents enabled in-depth investigation that helped to clarify ambiguities and guide subsequent work.
Together, these capabilities suggest a practical role for \name in other large datasets such as Perturb-seq atlases, where the hypothesis space is too broad to evaluate manually.

\hspace*{1.5em}The audit agents also exposed important failure modes.
For example, post hoc auditing revealed that \name had defined paralogs in Round 3 using gene-symbol prefixes, producing biologically incoherent pseudo-paralog families (e.g. classifying as paralogs all genes starting with \textit{CD}) and compromising the conclusion of the round.
More broadly, \name often jumped to running large numbers of statistical tests and emphasized results passing stringent filters, as opposed to presenting analyses as exploratory.
This required the scientist to examine data beyond reported significance to assess the analyses' value in steering hypotheses.
Initial hypotheses also often required substantial editing before they posed biologically meaningful questions.
These errors show that scalable hypothesis testing does not remove the need for expert oversight.
Experts have a crucial role in setting high-level direction, checking biological assumptions, and interrogating consequential analyses.
\end{steeringobservations}

\clearpage
\section{Discussion}
\label{sec:discussion}
{\spacing{0.93}Here, we present \name, a scientist-centric agentic AI system designed to support long-horizon discovery through repeated cycles of autonomous investigation and scientist-directed hypothesis steering. 
Across two predictive modeling tasks, \name produced models that outperformed both baseline agentic systems and many prior human-engineered models, including the highest performing model on the StarEmbed dataset and, with scientist steering, a variant-effect predictor that ranked second overall, behind only one of 106 reference predictors and above all comparable AI co-scientist agents.
Across three case studies in extragalactic astronomy, economics, and functional genomics, \name supported open-ended investigations that produced findings with direct relevance to their respective fields.
Taken together, these results suggest that a scientist-centric agentic system such as \name can contribute to meaningful scientific advancements across a broad swath of fields.

Across the scientist-steered investigations, we observed a consistent division of labor between broad empirical execution and scientific judgment. 
\name implemented analyses, tested modeling strategies, and traversed large hypothesis spaces, while scientists reframed objectives, corrected domain-specific assumptions, standardized evaluation procedures, interpreted ambiguous findings, and decided when a line of inquiry warranted further investment. 
These interventions occurred at epistemically consequential moments rather than at every computational step.
\name retained autonomy within a scoped hypothesis branch while the scientist directed how the investigation evolved across branches. 
The preservation of negative results in the accumulating research record also allowed failed approaches to narrow the search space and inform subsequent hypotheses rather than disappearing from the final account. 
This structure offers a practical middle ground between continuously supervising individual analyses and delegating an entire investigation to an autonomous system.

However, meaningful hypothesis steering depends on whether the scientist can understand the evidence accumulated before each decision. 
In \name, checkpointing makes the research process transparent while audit agents make the research record operational by locating supporting evidence, reconstructing analyses, and testing claims through follow-up work. 
Transparency makes a trace available, auditability makes it interrogable, and neither alone guarantees that a scientific conclusion is correct. 
Audit agents identified corrupted claims with an F1 score of 0.965 and located a test-set access violation precisely enough to restore an earlier checkpoint and resume the investigation under stronger controls. In a preliminary study, reviewers with audit access recovered a mean of 3.50 distinct, deliberately inserted errors per person, compared with 0.25 among reviewers given the paper and code.
Auditability functions not only for post hoc verification, but as a prerequisite for informed steering, allowing scientists to correct, qualify, or redirect research while preserving the useful work that preceded an error.

A number of limitations exist with both the \name system and our study presented here. 
First, \name is an AI system and suffers from all drawbacks related to AI systems, such as misrepresentation risk and hallucination. 
Second, \name is built solely for computational analyses and is not equipped for physical laboratory experiments.
Third, \name is an expensive system that is reliant on LLM and computational resource access, which might be cost-prohibitive for some scientists.
Because of this last point, we acknowledge a few drawbacks to our study design.
We do not provide comprehensive ablations on various components of \name due to the large amount of resources required to carry out these tasks.
In addition, we are limited to a relatively small sample size of scientists interacting with \name, due to the time and cost of running an end-to-end investigation.
Systematic studies of these components are left for future work, and we invite the community to rigorously test \name on a variety of scientific problems.

Evaluating AI scientist systems remains challenging because the value of scientific research cannot be fully represented by a single quantitative metric. 
We therefore evaluate capabilities of \name at three levels. 
First, held-out performance on StarEmbed and ProteinGym provides a direct quantitative comparison of the predictive models produced by different agentic systems. 
In the ProteinGym study, DA-Steer performed better than the evaluated autonomous agents, although differences in execution models and research budgets prevent attributing this advantage solely to scientist steering.
Second, the automated and human-assisted audit experiments test whether the research process is organized in a form that supports claim verification and error identification. 
Third, the three open-ended case studies provide empirical evidence that \name can participate in scientifically meaningful investigations for which research quality depends on domain-specific interpretation as well as numerical performance.
These evaluations do not constitute a comprehensive measure of scientific discovery or a universal ranking of AI scientist systems; rather, they triangulate predictive performance, auditability, and practical scientific use across settings in which no single metric captures the full research contribution.

Future versions of \name could extend this interaction model to parallel hypothesis exploration, richer multi-agent critique, and collaboration amongst interdisciplinary research teams. 
The central challenge in these scaling directions will be aggregating a growing body of agent-generated evidence without overwhelming scientists or obscuring the provenance, uncertainty, and contradictions that should guide the next decision. 
Preserving auditable evidence and deliberate hypothesis boundaries as the system scales may allow scientists to delegate increasingly broad empirical exploration while retaining authority over the questions pursued and the standards by which evidence is judged. 
More broadly, \name provides a framework in which agentic autonomy and scientist agency reinforce one another, supporting a future of AI-assisted science that remains interactive, inspectable, and directed by scientific judgment.
\par}

\clearpage

\section{Methods}
\label{sec:methods}
\subsection{System overview}
\label{sec:methods:overview}

\name takes a dataset and a research brief as input and returns a structured scientific report together with a complete, auditable research record.
The system is built from two nested loops.
An \textit{inner loop}, driven by a \textit{director agent}, orchestrates a set of tool agents---a code-executing agent, a web-search agent, a deep-research agent, and optional domain-specific agents---to carry out the analyses that test a single hypothesis.
An \textit{outer loop}, driven by a \textit{hypothesis brancher}, seeds each inner-loop invocation with a hypothesis, accumulates findings across invocations in a dedicated memory, and exposes points at which a human scientist can intervene.

\name is designed to span a spectrum of autonomy rather than a single operating point.
Run end-to-end in autonomous mode, it generates and selects its own hypotheses, sequences its own tool calls, and iterates until its budget is exhausted.
In steering mode, a scientist can step into individual control points---selecting a hypothesis, directing an analysis, or auditing a finished run---without changing the rest of the pipeline.
Audit agents and checkpoints fit naturally into this process, allowing scientists to rigorously evaluate \name outputs.

We first describe the agentic architecture in terms of the inner loop, which includes the primary tooling of the system, and the outer loop, which includes the hypothesis branches and long-range memory.
We then describe how human scientists interact with the system, including the hypothesis steering.
After this, we walk through the checkpointing mechanisms and how this enables audit agents.
Finally, we describe details for all experiments in the Results.
We defer many low-level details (thresholds, prompts, and file layouts) to the Supplementary Notes, Figures, and Tables.

\subsection{Inner loop: the director agent}
\label{sec:methods:director}

The core of the inner loop is the \textit{director agent}, an LLM prompted with the dataset description, the current hypothesis, a catalog of available tool agents, and the report accumulated so far in the round.
It performs the high-level control of testing the hypothesis.
The director's main job is to dispatch queries to \textit{tool agents}.
A \textit{query} is the unit of the inner loop as it represents one call to a tool agent from the director.
A contiguous sequence of queries that investigates one hypothesis is a \textit{round}, and a full run consists of a configurable number of rounds.

The unit of the inner loop is a \textit{query}, which is a single message the director dispatches to one tool agent.
Two budgets govern a run and are set independently: a per-round query budget controls how deeply each hypothesis is pursued, and a round budget controls how many hypotheses are pursued in sequence.
We discuss round budgets in a later section (Methods Sec.~\ref{sec:methods:outerloop}).
The director is prompted to output a ReAct-style~\cite{yao2023react} natural-language rationale followed by a single tool agent invocation, written as an XML-style tag \texttt{<agent>query</agent>}, where \texttt{agent} is the name of the tool agent and \texttt{query} is the text of the query passed to that sub-agent.
Upon receipt of the query $q_i$, the tool agent $\theta_j$ returns to the director agent a natural-language report $\mathcal{R}_i$, which is the corresponding report for the round. 
This report is injected into the director agent's conversation and used as context for the next query.
The director runs as a persistent multi-turn conversation with sub-agents for the length of a round and invokes exactly one tool agent per query.
This recurrent director--tool-agent exchange defines the inner-loop information flow (Supplementary Figure~\ref*{si:fig:agent_innerloop}).

\xhdr{Director as the central controller}
The director, given a hypothesis and optional human guidance (described in Sec.~\ref{sec:methods:hitl}), decides the shape of a round.
This flexibility allows the director to repeat failed queries, retry queries where results may be ambiguous, or to stop a round early if the director has enough evidence to validate or invalidate the hypothesis.  
When enabled, we prompt the director to emit the reserved \texttt{<<STOP>>} output once it judges the hypothesis sufficiently explored; the workflow detects this signal, terminates the round before exhausting its query budget, and proceeds to report generation.
Through prompting, we restrict the director to executing only analyses explicitly related to the round's hypothesis and guidance, preventing diverging inquiries that were not requested by the hypothesis selector or human driving the research process.
This gives scientists more explicit control over the research process.
These behaviors distinguish \name from agents that execute a predetermined sequence of steps; the director decides when an analysis is complete, when to repeat work, and when to move on.
Since the director agent is a crucial point for success of \name, we often use a larger and more performant LLM for this component, such as Claude Opus-level models (see Sec.~\ref{sec:methods:configurations} for details). 
The per-component model assignments and the round and query budgets used in each experiment are given in Supplementary Note~\ref*{si:note:model_config} and Supplementary Table~\ref*{si:tab:model_config}.

\xhdr{Context management}
Long rounds with verbose tool outputs can exceed the director's context window.
We perform context isolation on the director-side through two mechanisms.
First, the report structure compresses sub-agent work into one central natural-language description of the research, preventing context bloat for the director agent for routine tool calls.
Second, the director compacts its own context. 
Once its estimated input length crosses a threshold, earlier turns are replaced by a model-generated summary that preserves the hypothesis, the results obtained so far, and the outstanding next steps.
The same mechanism is triggered as a fallback when a context-overflow error is returned mid-query, so that rounds with large per-query outputs continue without manual intervention.
When compaction is unavailable, the query budget bounds the conversation length.
Supplementary Note~\ref*{si:note:compaction} gives the provider requirements, the trigger thresholds, and the information the summary is required to retain.
Note that this context management is separate from the memory, which operates on the outer loop, discussed in Methods Sec.~\ref{sec:methods:memory}.

\subsection{Inner loop: tool agents}
\label{sec:methods:tools}

The director dispatches work to a set of tool agents, each exposing a narrow interface and registered in the director's catalog with a short description of when to use it and what it returns.
Three general-purpose tool agents are available in every run.

\xhdr{Web-search agent}
The web-search agent performs a single-shot search using the underlying model's native web-search capability and returns a short text answer.
It is meant for lightweight lookups---confirming a definition, locating a canonical reference, or finding a publicly reported statistic---that do not justify the cost of a deep research query.

\xhdr{Deep-research agent}
The deep-research agent issues a long-form query to a dedicated deep-research model and returns a structured report with an executive summary, key findings, a methodology note, limitations, and a list of sources.
It is used for questions that require synthesis across many sources, such as surveying prior approaches to a task or summarizing the literature around a candidate mechanism.
We use Perplexity's \texttt{sonar-deep-research} model throughout our experiments, but the \name system supports other models as well.

Before a run, \name can seed hypothesis generation with a literature background report.
When no reference collection is supplied, the deep-research agent uses the dataset description and research brief to synthesize relevant prior work, established methods, and candidate research questions; alternatively, a scientist can supply a collection of literature, which \name summarizes and synthesizes without drawing on outside sources.
In both modes, \name converts the literature into a single report that is provided as context during hypothesis generation rather than inserting the full source documents into agent conversations.
The director can subsequently invoke the deep-research agent for targeted questions that arise during hypothesis testing.

\xhdr{Execution agent}
The execution agent handles the computational work such as loading data, running statistical tests, fitting models, generating figures, and following up on intermediate results.
It is an autonomous coding agent built on the \texttt{claude\_agent\_sdk}, with standard file and shell tools inside a sandboxed working directory~\cite{anthropic_agent_sdk}.
Each invocation is a self-contained session where the agent receives a query from the director, writes and runs code, saves figures and intermediate artifacts to the working directory, and returns only a concise summary of what it did and found.
Returning the summary rather than the full tool-call trace keeps the director's context focused on scientific content rather than implementation detail.
To reduce accidental access to files and benchmark results outside the research workspace, we use layered application-level containment where pre-tool hooks constrain direct file operations and screen common shell access patterns, a post-tool guard terminates sessions after detected out-of-sandbox writes or edits, and a targeted URL hook blocks known ProteinGym result sources.  
Supplementary Note~\ref*{si:note:execution_agent} describes the execution environment and summary generation, while Supplementary Note~\ref*{si:note:leakage_audit} details the containment hooks, the ProteinGym access event, and the subsequent hardening and rollback.

\xhdr{Director agent orchestration of tool agents} 
The director agent acts as the central intermediary between all the tool agents, passing relevant findings between sub-agents to facilitate the testing of the hypothesis.
Within individual hypothesis-testing rounds, the director generally followed a primary test $\rightarrow$ diagnosis $\rightarrow$ challenge $\rightarrow$ validation sequence.
It first operationalized the central prediction and requested a direct comparison, then used the result to select targeted tests of confounding, alternative explanations, subgroup or condition dependence, and robustness.
For example, in the functional genomics study, an unexpected enrichment of primary immunodeficiency genes among relay genes led the director to request logistic regression, expression adjustment, condition stratification, overlap exclusion, permutation testing, and in-degree-matched controls. 
In the photometric-redshift prediction study in extragalactic astronomy, an initial performance improvement accompanied by rapid overfitting prompted successive tests of regularization, backbone freezing, alternative architectures, boosted trees, and ensembles.
Specialized agents typically entered this sequence to resolve a specific information bottleneck---such as acquiring Census variables in the economics study or disease annotations from OpenTargets in the functional genomics study---after which the execution agent integrated the new evidence and performed the hypothesis test.
Across the three case studies, the execution agent accounted for 87.5--94.6\% of classifiable within-round calls, while web search and domain-specific agents were used selectively when a test required external evidence (Supplementary Note~\ref*{si:note:resource_accounting}).
Deep research agents were used sparingly by the director agent, with the only deep research usage arising in the DA-Steer trial of the ProteinGym experiments.
The web search agents were more often used to gather quick references or more precise queries.
This was most likely due to either the seeding of deep research reports in each run or the scientist-supplied hypotheses often already including relevant literature details, making comprehensive deep research queries less useful in the context of the round-wise budgets.

\subsection{Outer loop: hypothesis branching}
\label{sec:methods:outerloop}

The outer loop wraps the inner-loop workflow and runs it over a configurable number of rounds.
Each round proceeds in three phases: a hypothesis is selected and used to condition the director (defined as \textit{hypothesis branches}); the inner loop runs until it terminates or exhausts its query budget; and the resulting report is folded into the accumulated memory.
Between rounds the in-round conversation is reset, so each round starts from a clean director context while the accumulated memory persists and is re-injected as background at the start of the next round.
In this section, we'll describe the overview of the outer loop, and subsequent sections will outline the hypothesis generation.

\name can take on two modes based on outer loop autonomy.
In \textit{automatic} mode, a hypothesis-generation agent proposes a single hypothesis---a statement with a justification grounded in the literature or in the accumulated memory, and an optional plan for testing it---which is directly passed to the director.
In \textit{hypothesis-steering} mode, described in later sections, the generated hypotheses are handed to a human scientist, giving the scientist agency to edit, select, and refine hypotheses before choosing to run them in the inner loop.
Optionally, the first few rounds can be run in an \textit{exploratory} mode in which no hypothesis is fixed and the director is instead asked to broadly characterize the dataset, seeding later hypothesis generation with observations from the data rather than from the literature alone.
Supplementary Note~\ref*{si:note:model_config} reports the round, branch, exploratory-round, and per-round query settings used in each experiment, while Supplementary Note~\ref*{si:note:checkpointing} specifies the candidate, selected-hypothesis, round-report, and accumulated-memory artifacts that connect successive rounds.

\subsection{Memory: empirical and procedural}
\label{sec:methods:memory}

\name keeps two separate kinds of long-lived state across rounds: an \textit{empirical memory} of findings and a \textit{procedural memory} of code.
Our central intuition behind this separation is that the director agents need only understand what has been \textit{found} (empirical memory), and the execution agent needs to understand what has been \textit{implemented}.

\xhdr{Empirical memory}
The empirical memory is an agentic memory that integrates each new round's report while pruning redundancy, using a three-agent pipeline adapted from the ACE context-engineering architecture~\cite{zhang2025agentic}.
An \textit{integrator} merges the accumulated memory with the new report with completeness as its explicit objective, retaining every data point and technical detail from both.
An \textit{analyzer} then identifies passages that could be consolidated and assigns each a risk level (low, medium, or high) based on the technical detail that consolidation might remove, while specifying the information that must be preserved.
A \textit{refiner} applies only the low-risk consolidations, verifying that information flagged as critical is preserved.
Similar to the motivation behind ACE, splitting the update into three narrowly scoped agents avoids dropping details and making overgeneralizations that result when a single agent is asked to integrate, analyze, and rewrite at once.
This memory is injected into the director and the hypothesis-generation agent at the start of each round.
This is how \name accumulates findings across rounds without ever passing a full per-round conversation forward.
Supplementary Note~\ref*{si:note:empirical_memory} details the pipeline.

\xhdr{Procedural memory}
The empirical memory is crucial for long-range understanding of research results, but it is not built to hold low-level details about codebase construction.
Since execution agents are launched with fresh context at each query in which the director agent invokes them, we use a procedural memory to track what has previously been implemented.
We use an active memory component and a passive memory component, both implemented via markdown files in the codebase. 
The active memory component is a short markdown file to orient the execution agent to the codebase when launching a fresh agent into the codebase.
The passive memory component is a longer documentation file that can be accessed on-demand with tools such as grep.
We build these memory components through two sub-agents that maintain each document; these sub-agents are managed by the central execution agent, which is prompted to maintain documentation of its code to increase codebase reusability.
We prompt these sub-agents to 1) keep the active memory as an overview of the codebase and 2) to make the passive memory comprehensive but \textit{greppable}, which is a useful framework to enable more efficient coding agent use of the file~\cite{yang2024swe}.
Supplementary Note~\ref*{si:note:procedural_memory} gives the file names and the sub-agent workflow.
We enforce some additional structure on the codebase structure to ensure that the execution agent grows the codebase in a reusable and modular way, borrowing proper software engineering principles.

\subsection{Review mechanism}
\label{sec:methods:review}

\name injects an automated reviewer into the research loop at a configurable query frequency.
When a review fires, the round's cumulative report is passed to a reviewer LLM that produces a critique focused on novelty, rigor, and direction; the critique is fed back to the director as feedback that shapes the next query.
During \textit{exploratory rounds}, we omit the reviewer in order to allow \name to explore the dataset without critique overhead; when executing specific hypotheses, we enable the reviewer.
The reviewer can optionally be granted web-search access to check whether a finding has already been reported in the literature.

\xhdr{Review designer}
Rather than a single fixed reviewer prompt across all tasks, \name uses a \textit{review designer} that generates a task-specific reviewer prompt at the start of each run from the dataset description and the research task. 
For example, the reviewer can emphasize causal rigor for an epidemiology task, or novelty for a mechanism-discovery task.
This ties the reviewer to the research goal rather than to a generic notion of scientific quality, and removes the need to hand-write a reviewer prompt for each run.
The generated prompt is checkpointed with the rest of the run configuration so the reviewer's behavior can be audited.
The experiment-specific review intervals and reviewer model assignments are reported in Supplementary Note~\ref*{si:note:model_config}; Supplementary Note~\ref*{si:note:checkpointing} describes how the active reviewer prompt and review-relevant workflow state are included in each query checkpoint.

\subsection{Steerability: Hypothesis Steering}
\label{sec:methods:hitl}

\xhdr{Scientist intervention at hypothesis boundaries}
In hypothesis-steering, we concentrate scientist intervention at the boundaries between research rounds (Figure~\ref{fig:overview}b). 
After each round, \name preserves the latest round report, accumulated empirical memory, checkpoint trace, and research codebase, then returns control to the scientist before beginning the next branch. 
The steering interface supplies the original research task, accumulated memory, and latest round report to a \textit{hypothesis generation agent}, which proposes a configurable set of candidate directions grounded in the evidence accumulated so far.
The checkpoint trace and codebase remain available for separate manual inspection by the scientist. 
Each candidate comprises a testable hypothesis, a justification, and an analysis plan; the scientist can additionally provide initial guidance to constrain the candidate set toward a scientific question, mechanism, or methodological priority.

\xhdr{Evidence-grounded selection and revision of hypotheses}
To support an informed choice rather than a review of candidate text alone, the interface exposes the accumulated memory and latest round report alongside conversational agents that can answer questions about the reports, checkpoint trace, and codebase. 
The scientist can therefore interrogate the evidence and assumptions underlying a proposed direction before 1) accepting a candidate, 2) editing it directly, 3) asking the hypothesis-generation agent to revise or regenerate it from written feedback, or 4) replacing it with a custom hypothesis. 
The hypothesis selection process is not a passive multiple-choice selection process but rather an active engagement from the scientist, allowing for rich internal feedback through a conversational format with the hypothesis generation agent.
In addition, the scientist has access to the audit agents, which allow for in-depth searching over the research checkpoints and codebase, allowing for deeper understanding of the scientific process to inform selection.
The scientist may also attach overall research guidance that applies beyond the hypothesis itself, such as a requirement to avoid causal language, use finer-grained data, or prioritize a particular validation strategy. 

\xhdr{Handoff to autonomous execution}
Finalization produces a single selected hypothesis artifact containing the hypothesis itself, its justification, an optional analysis plan, and any scientist feedback or overall guidance. 
When the run resumes, \name loads this artifact and injects it into the director's first query context with an explicit instruction to test the scientist-supplied hypothesis and integrate the accompanying feedback; the selected direction then remains in the director's conversation throughout the round. 
The director autonomously decomposes this direction into queries and executes the inner loop under the existing query budget, while the accumulated empirical memory and version-controlled codebase persist across rounds. 
Hypothesis steering thus adds a low-frequency, high-leverage intervention at the point where the research objective changes, rather than requiring the scientist to supervise individual tool calls. 
Supplementary Note~\ref*{si:note:hypothesis_steering} describes the interface state, revision operations, saved artifacts, and consistency controls, and Supplementary Note~\ref*{si:note:checkpointing} details the checkpoint and resume machinery.

\subsection{Transparency: checkpointing and version control}
\label{sec:methods:provenance}

\name implements an extensive checkpointing mechanism that logs each query and round action.
The checkpoint store is organized by round and by query: each query has its own JSON checkpoint of the full workflow state---the director and hypothesis-agent conversations, the intermediate results, the running report, any active reviewer prompt, and the run configuration---together with a human-readable Markdown rendering of the query and its result.
Each round additionally writes its starting configuration, its report before and after memory integration, its log, any generated reviewer prompt, and---where applicable---the selected hypothesis and the candidate list used to seed it.

Separately, the execution agent's working code directory is placed under Git~\cite{git} version control, and changes are committed as the run proceeds with messages that tag each commit by round and query.
This yields a navigable history of the code, data, and figures produced during a run, so that a reviewer can inspect the working directory at any point in the trace and diff between steps.
Together, the checkpoint tree (the state of the agents) and the git history (the state of the code and data) record both what the agents were reasoning over and what was on disk at every step.
Supplementary Note~\ref*{si:note:checkpointing} describes the store layout and the commit scheme.

\xhdr{Checkpoint resumption}
At startup, the outer loop scans the checkpoint tree to distinguish completed rounds from an incomplete next round. A completed round advances the run to the next hypothesis boundary with its accumulated memory and, in steering mode, the selected-hypothesis artifact; an incomplete round instead restores the most recent query checkpoint, including the running report, agent conversations, reviewer state, and query counter. This separation allows a scientist to pause deliberately at a hypothesis boundary while also allowing interrupted tool-agent work to resume without treating it as a new scientific branch.

\subsection{Auditability: Audit agents}
\label{sec:methods:audit_agents}

\name exposes the trace (checkpoints, memory files, and version control) to a set of \textit{audit agents}.
An audit pipeline consumes the per-query checkpoints and renders a run as a navigable, round-by-round timeline of queries, results, hypotheses, reviews, and code changes.
Audit agents are practically implemented with the Claude Agent SDK harness, which allows for access like grep and progressive context disclosure~\cite{anthropic_agent_sdk}.
However, audit agents could feasibly admit any harness, allowing for flexibility by downstream users; indeed, we use audit agents with the Codex~\cite{openai_codex_cli} harness in audit experiments (Section \ref{sec:results:audit}).
The construction of \name's checkpoints is built specifically to enable navigation towards detailed round-by-round, query-by-query transcripts, leveraging progressive context disclosure for efficiency and completeness.
We demonstrate the efficacy of this construction, as well as the power of audit agents equipped with \name's provenance structure, in Section \ref{sec:results:audit}.

Oriented with these components, audit agents operate in two distinct configurations that differ in what they are permitted to do to the trace.
The \textit{read-only} configuration observes the research from outside it. 
Its tool allow-list contains only file reading, content search, globbing, and shell commands, with no write, edit, or web-fetch capability, so a session cannot alter or rerun the research under review.
This is the configuration exposed through the audit interface during hypothesis steering (Supplementary Note~\ref*{si:note:audit_ui}), given to the Audit cohort of the human study, and used as the fixed auditor in the corrupted-claim benchmark; it answers questions and traces the provenance of an implementation or finding, but every claim it makes is grounded in artifacts that already exist.
The \textit{execution-enabled} configuration additionally receives write and execute permissions so that it can reproduce a reported result or run a follow-up analysis.
It mounts the checkpoint tree and research codebase read-only and writes into a separate analysis directory, so the original \name implementation is preserved and any new script is distinguishable from agent-generated research code.
We used this configuration for the post hoc reproductions and extensions reported in the case studies, including the economics premium-trend and price-to-rent analyses and the functional genomics topology, tractability, and target-prioritization follow-ups; the reproduction scripts are named in the corresponding Supplementary Notes.
Read-only auditing is therefore the configuration that supports steering and review, whereas execution-enabled auditing is what produces new analysis on top of a completed run.

Audit agents are a simple yet powerful framework, and the \name system is built to natively integrate them into both hypothesis steering and follow-up analyses.

\subsection{Specialized agents}
\label{sec:methods:specialized}

The modularity of the tool agents allows \name to integrate specialized agents that can integrate domain-specific tooling.
We use specialized tool agents in two of our case studies to expose curated access to specific data sources and domains.
These agents follow the same interface as the general-purpose agents but are enabled only on runs where their data source is relevant; we describe the two used in this work below.

\xhdr{Census data agent}
The economics experiment is served by the census data agent, a data-delivery agent that exposes three public-use data backends through a single search-then-download interface: published American Community Survey (ACS) summary tables, confidentiality-protected ACS Public Use Microdata Sample (PUMS) microdata, and harmonized IPUMS microdata (the USA, CPS, NHIS, and MEPS collections).
Queries are phrased in terms of geography (from nation down to block group and Public Use Microdata Area, plus metropolitan areas, congressional and school districts, and ZIP-code tabulation areas) and subject matter; each download returns tabular files of estimates---with optional margins of error---accompanied by a JSON data dictionary explaining every variable code. 
The agent populates an initially empty data directory on demand and is restricted to data retrieval. 
We document the surveys, tables, geographies, and variables accessible through these backends in Supplementary Note~\ref*{si:note:census_data} and Supplementary Table~\ref*{si:tab:acs_table_series}. The Supplementary Note provides the exact table-, release-, and file-level citations used in this study, including the required Census Bureau and IPUMS attributions.

\xhdr{OpenTargets agent}
The OpenTargets agent exposes the Open Targets Platform's curated target--disease association data, stored as parquet files in a shared read-only directory.
The agent spawns a sandboxed coding session that reads from this directory and writes tabular extracts (CSV or parquet) for downstream analysis; a write-protection hook prevents it from modifying the shared source, and a process-control hook blocks long-running or destructive commands.
Its prompt documents the major entity classes of the Open Targets schema (genetics, disease, association, target, expression, and drug).
Supplementary Note~\ref*{si:note:opentargets_agent} catalogs the complete mounted release, both query interfaces, and the sandbox controls.

\subsection{Shared agent configurations}
\label{sec:methods:configurations}

All experiments use the architecture described above while varying the operating mode, component models, round and query budgets, review frequency, context compaction, early stopping, and availability of specialized agents. StarEmbed and the autonomous protein variant-effect prediction run (DA-Auto) use automatic hypothesis selection, whereas the extragalactic astronomy, economics, functional genomics, and scientist-steered protein experiment (DA-Steer) allow a scientist to select or edit hypotheses between rounds. Supplementary Note~\ref*{si:note:model_config} reports the complete component assignments and budgets for each experiment.

Across the five scientific studies, the director, hypothesis, review, review-design, memory, and web-search components were configured with Claude Opus~4.6, while deep-research calls use Perplexity's flagship agent \texttt{sonar-deep-research}. 
Execution models differ by study: StarEmbed uses Claude Sonnet~4.5, the ProteinGym configuration uses Sonnet~4.6, the extragalactic astronomy and economics studies use Opus~4.6, and the functional genomics campaign changes execution models during the trace. Query caps are 10 per round for StarEmbed, ProteinGym, and extragalactic astronomy, and 25 for economics and functional genomics; available assignments and deviations are listed in Supplementary Note~\ref*{si:note:model_config}.
Longer query numbers are given to the economics and functional genomics experiments because these experiments required iterative interaction with a data acquisition specialized sub-agent, which may require longer rounds.
For literature/deep-research seeding, we supplied a small collection of initial papers to the extragalactic astronomy experiment, whereas the other four studies began with an automatically generated deep-research background report.

\subsection{Predictive modeling evaluation}
\label{sec:methods:predictive_evaluation}

We separated model development from final assessment in predictive modeling evaluations for StarEmbed and ProteinGym. 
Before exploration, we fixed the test split and provided each agent with only the labeled research data.
For StarEmbed, the research data included a preselected validation set; for ProteinGym, it contained only the training variants, and each agent selected its own internal validation data.
Agents could use validation data throughout exploration to compare hypotheses, features, model classes, and hyperparameters, but no part of the test set was exposed in the research workspace.
To separate these sets, we define these as the \textit{research split}, or the data given to the agent to perform the research, and the \textit{evaluation split}, the data used to evaluate the agent.
We specified task-appropriate reporting metrics in advance and used the same metric definitions for validation and final testing; validation metrics guided development, although individual training procedures could optimize auxiliary objectives such as multiclass log loss, whereas test metrics were reserved for final performance reporting.
In the extragalactic astronomy case study, we retrain and evaluate selected recipes on a separately constructed sample from the same survey sources (Supplementary Note~\ref*{si:note:dwarf_dataset}). Its training, validation, and test splits are mutually disjoint, but $26.7\%$ of its test galaxies also occur in the original research training or validation data due to overlaps in source data; Supplementary Note~\ref*{si:note:dwarf_overlap} reports sensitivity analyses excluding these galaxies and shows that excluding these galaxies does not meaningfully change performance.

After exploration concluded, we mechanistically identified the best model reported by each agent, without using test-set performance to choose among candidates. 
We then extracted the corresponding training and evaluation code from the agent's research artifacts, including its preprocessing, feature construction, model configuration, and metric implementation. This procedure ties each reported result to the representation and model class selected within the original research process.
Supplementary Notes~\ref*{si:note:starembed_da_model}, \ref*{si:note:starembed_classifier}, and~\ref*{si:note:baseline_agents_dms} document implementation-specific refitting and stored-prediction decisions.

Once the selected representation, model class, and final training procedure were fixed, we retrained on the research split where refitting was required, and evaluated on the untouched test set. 
We report the resulting prespecified test metrics as final model performance.
Supplementary Notes~\ref*{si:note:starembed_agent_baselines} and~\ref*{si:note:starembed_classifier} describe the StarEmbed data release, selected artifacts, metrics, and held-out evaluation. 
Supplementary Notes~\ref*{si:note:baseline_agents_dms}, \ref*{si:note:dms_auto_architecture}, and~\ref*{si:note:dms_steer_architecture} provide the corresponding details for the ProteinGym case study.

\subsection{Setups for predictive experiments}
\label{sec:methods:experiment_setups}
We reproduce the dataset- and task-description seed inputs for the predictive studies and scientist-steered case studies in Supplementary Prompt Notes~\ref*{si:prompt:campaign_seeds}--\ref*{si:prompt:campaign_seeds_last}.

\xhdr{StarEmbed setup} We ask \name to develop a seven-class variable-star classifier using the StarEmbed research split, which contains 29,047 training and 4,150 validation objects. 
Each object includes irregularly sampled light curves in up to three bands ($g$, $i$, and $r$), with observation times, magnitudes, measurement uncertainties, a catalog period, and a class label. 
We supply a description of the data schema but leave feature construction and classifier design to the agent.
We use automatic hypothesis selection with Claude Opus~4.6 as the director and Sonnet~4.5 as the execution agent, with director context compaction enabled and without specialized data agents.
The run completes one exploratory round followed by three hypothesis-driven rounds, totaling 38 tool-agent queries under a ten-query cap per round. 
For final evaluation, we retain 7,955 of the original 8,301 test samples with more than ten observations in both $g$ and $r$, ensuring sufficient coverage for the light-curve features. 
Supplementary Notes~\ref*{si:note:starembed_trace} and~\ref*{si:note:starembed_classifier} describe the research trajectory, selected classifier, and evaluation protocol.

\xhdr{ProteinGym setup} We ask \name to predict the functional effects of amino-acid substitutions using 555,501 training variants from 217 ProteinGym deep mutational scanning assays. 
The research bundle includes mutated sequences and measured effect scores, an assay index linking reference sequences to associated files, 194 multiple-sequence alignments, and 197 AlphaFold2 structures. 
We compare automatic hypothesis selection (DA-Auto) with scientist steering between rounds (DA-Steer), using the same input bundle and Claude Opus~4.6 director in both settings. DA-Auto uses Sonnet~4.6 for execution, whereas DA-Steer uses Opus~4.6 for Rounds 0--7 and Sonnet~4.6 for Rounds 8--10 and the subsequent aggregation step; both runs enable director context compaction and use the general-purpose tool agents.
DA-Steer was configured with 11 rounds, including one exploratory round and ten scientist-intervention rounds, followed by a separate aggregation step.
The per-round query cap is ten, increased to fifteen for DA-Steer Round~10 and the subsequent aggregation step.
We evaluate the resulting predictors on 140,810 held-out variants from non-overlapping residue positions within the same assays, reporting per-assay Spearman correlation. 
The evaluation split contained contiguous region holdouts, ensuring strong generalization requirements that avoided memorizing local structural patterns for the protein.
We score all 106 ProteinGym reference predictors on these same held-out variants, using ProteinGym's out-of-fold predictions for the supervised references (Supplementary Note~\ref*{si:note:baseline_agents_dms}).
Supplementary Notes~\ref*{si:note:dms_auto_trace} and~\ref*{si:note:dms_steer_trace} give the hypothesis tested in every round. Supplementary Notes~\ref*{si:note:dms_auto_architecture} and~\ref*{si:note:dms_steer_architecture} specify the final predictors, and Supplementary Note~\ref*{si:note:leakage_audit} documents the benchmark-access violation and rollback in DA-Auto.

\subsection{\texorpdfstring{Baseline agents}{Baseline agents}}
\label{sec:methods:baselines}

We compare \name with four agentic baselines spanning general-purpose coding, parallel multi-agent investigation, iterative hypothesis testing, and agentic tree search; task inputs, budgets, models where exposed, artifact-selection rules, and held-out evaluation procedures are reported in the Supplementary Information.
For each benchmark, the baseline agents received the same released research split; system-specific setup and prompting are documented in Supplementary Notes~\ref*{si:note:starembed_agent_baselines} and~\ref*{si:note:baseline_agents_dms}.
We exclude methods designed for strictly verifiable tasks because \name does not assume an external verifier and the held-out test sets remain unavailable during research.

\xhdr{Coding agent} We evaluate a general-purpose coding agent in single-session (referred to as ``Coding agent@1'') and fixed-goal continuation settings (referred to as ``Coding agent loop'').
Coding agent@1 received one prompt per task, with experiment-specific turn and scheduler caps, while the Coding agent loop received the initial prompt followed by a fixed number of continuation prompts.
Experiment-specific configurations and artifacts are reported in Supplementary Notes~\ref*{si:note:starembed_agent_baselines}, \ref*{si:note:starembed_classifier}, and~\ref*{si:note:baseline_agents_dms}.

\xhdr{Kosmos} Kosmos is an autonomous AI scientist system that decomposes an open-ended research objective into parallel investigations and integrates their findings into a report~\cite{mitchener2025kosmos}.
We use Kosmos through Edison's provided online interface~\cite{edison_kosmos_platform}. 
Prompts, hosted-interface settings, artifacts, and evaluation decisions are reported in Supplementary Notes~\ref*{si:note:starembed_agent_baselines}, \ref*{si:note:starembed_classifier}, \ref*{si:note:baseline_agents_dms}, and~\ref*{si:note:kosmos_exclusion}.

\xhdr{AutoDiscovery} AutoDiscovery is another autonomous AI scientist system that iteratively proposes and tests hypotheses using Bayesian surprise~\cite{agarwal2026autodiscovery}.
Similar to Kosmos, we use the online interface for AutoDiscovery for our tasks~\cite{ai2_autodiscovery_interface}.
Experiment budgets, prompts, returned hypotheses, and downstream evaluation are reported in Supplementary Notes~\ref*{si:note:starembed_agent_baselines}, \ref*{si:note:starembed_classifier}, and~\ref*{si:note:baseline_agents_dms}.

\xhdr{AI-Scientist-v2} AI-Scientist-v2 uses agentic tree search to implement and refine generated research ideas~\cite{lu2026towards}.
Since AI-Scientist-v2 was originally designed to source problem statements from literature instead of working on fixed datasets, we adapted AI-Scientist-v2 with a project-authored wrapper that supplied each benchmark's task description and local data to its native ideation and tree-search stages, omitted its paper-writing pipeline, and applied its reviewer separately to the stage summaries.
Because of AI-Scientist-v2's original design to gather datasets as part of its process, it refused to use the provided datasets during its analyses, despite being given access to the same datasets as the other baseline agents.
This resulted in the methods built by the agent often being built on synthetic data.
The prompts, search configurations, candidate-selection procedures, and evaluation provenance are reported in Supplementary Notes~\ref*{si:note:starembed_agent_baselines}, \ref*{si:note:starembed_classifier}, and~\ref*{si:note:baseline_agents_dms}.

\xhdr{Model selection and resource control} For selecting models from each agent, we followed closely our implementation of model selection from \name. 
We used the agent's own internal reporting to rank the best model to choose; this was based on either an explicit recommendation from the baseline agent or reported validation losses.
The model-selection decisions are documented in Supplementary Notes~\ref*{si:note:starembed_classifier} and~\ref*{si:note:baseline_agents_dms}.
We also controlled resources for each agent to the furthest configurable extent. 
Both coding agent implementations as well as the AI-Scientist-v2 implementations were given access to the same computational resources as \name for the tasks.
We were unable to provide this same control for Kosmos and AutoDiscovery as these baselines were run through the available online interfaces.
On agent budget, we also attempted to control the number of experiments for which each agent could implement models on the dataset.
This was very feasible for all agents except for Kosmos, which is given no lever to control experimentation effort through the online interface.
The baseline resource and experiment budgets are documented in Supplementary Notes~\ref*{si:note:starembed_agent_baselines} and~\ref*{si:note:baseline_agents_dms} and Supplementary Table~\ref*{si:tab:starembed_baseline_config}.
The allocations, realized budgets, and cost accounting for every locally executed run are given in Supplementary Note~\ref*{si:note:resource_accounting} and Supplementary Table~\ref*{si:tab:resource_accounting}.

\subsection{Evaluation of audit agents}
\label{sec:methods:audit_evaluation}

We evaluate audit agents in an automated benchmark measuring whether a fixed auditor can identify corrupted claims from research traces and in a human study measuring whether access to an audit agent improves expert review of an agent-generated paper.
The automated benchmark varies the trace while holding the auditor fixed, whereas the human study varies whether reviewers receive the read-only audit interface in addition to the paper and codebase. 
We score the automated benchmark mechanically from known corrupted claim identifiers; the human study requires a separate matching rubric for free-text reviewer findings. 

\xhdr{Automated corrupted-claim identification} We first extract 50 claims from each final report produced by \name, Kosmos, and the coding agent loop baseline on StarEmbed. 
For each system, we construct five independently sampled cases in which 25 claims are corrupted using a fixed catalog of eight corruption types.
The corruption catalog comprises numeric alteration, reframing, effect-size inflation, limitation removal, method misstatement, significance misstatement, validation misstatement, and sample-size alteration. 
Thus, a case may change a reported value or sample size, strengthen the interpretation of an effect, remove a stated caveat, or claim a validation procedure that the trace does not contain. 
We sample the corrupted claim identifiers with a fixed seed for each case, while the corruption type assigned to a selected claim follows the fixed generation protocol described in Supplementary Note~\ref*{si:note:audit_claim_corruption}.

We then feed these 50 claims from each agent, along with that respective agent's research artifacts (Supplementary Note~\ref*{si:note:audit_trace_provenance}), to coding agent auditors.
We evaluate GPT-5.6-Sol, GPT-5.6-Terra, and GPT-5.6-Luna with medium reasoning in the same Codex harness~\cite{openai_gpt56_system_card, openai_codex_cli}. Within each auditor-model comparison, we hold reasoning effort, tool access, and response schema fixed across trace sources.
We used GPT-5.6 in this case to differentiate from the Claude models used to insert corruptions.
The auditor labels every claim as supported, misrepresented, or uncertain, and we score only the misrepresented verdict as a positive detection against the known corruption set.
The scoring is done deterministically, using each ``corrupted'' claim number and matching against ground-truth; this avoids ambiguous LLM-as-a-judge adjudications. 
We also run a matched report-only control that removes access to the trace while holding the reports, corruptions, and auditor fixed.
Each run uses a fresh read-only environment with network access disabled and with memory, apps, plugins, and nested agents disabled. 
Supplementary Note~\ref*{si:note:audit_claim_corruption} gives the full construction and scoring protocol.

One possible explanation for the performance gap is that the inserted corruptions differed in difficulty across systems. To assess this, we obtained post hoc, system-blinded difficulty ratings from independent LLM graders and evaluated recall by difficulty stratum and with logistic adjustment, without reselecting or rebalancing the corruptions. Although difficulty distributions differed, \name retained higher recall within the shared difficulty strata and after adjustment (Supplementary Note~\ref*{si:note:audit_corruption_difficulty}, Supplementary Table~\ref*{si:tab:audit_recall_by_difficulty}).

\xhdr{Human-assisted error identification} We conduct an experiment to test whether audit agents help machine-learning researchers identify errors in an agent-generated research paper.
Using an intermediate run of the DA-Auto run for the DMS experiment (version with five completed rounds), we used audit agents to generate a report in the style of an academic paper.
We then used LLMs to identify areas of the paper that could be manipulated or misrepresented to insert corruptions.
Using LLMs, we identified 25 corruptions per paper, prompting the LLM to classify each error as ``easy'', ``medium'', or ``hard'' to find.
We repeated this process 3 times, yielding 75 total corruptions.
We first removed the ``easy'' corruptions as most of these involved obvious, large changes to the manuscript that would have been easily detectable.
We then manually sorted through each of the corruptions, removing duplicates and those that would be almost impossible to find in the timeframe such as slight numeric changes that are within error bars.
This left us with 36 errors; we then used an LLM to apply these changes to the paper.

We recruited 8 participants and randomly sorted them into either the \textit{code} group ($n=4$) or the \textit{audit} group ($n=4$). 
The code group received only the codebase and final corrupted report.
The audit group was given the codebase, the agent checkpoints, the corrupted final report, and the audit agents attached to each of these artifacts.
All participants received the same corrupted paper and codebase, holding the inserted errors fixed across cohorts.
The code group was explicitly instructed not to use AI tools, including chat interfaces, coding agents, or AI-enabled editors; standard code-search tools such as keyword search were permitted.
Further, we facilitated the audit group through an online user interface which gave them access to the audit agent for consistency.
Importantly, all auditors across both cohorts were given the same corrupted paper and codebase to control for strength of corruptions.
We then asked reviewers to identify inconsistencies or errors as in a conference review and to record a short summary, location, and explanation for each finding under a self-tracked one-hour limit (Supplementary Note~\ref*{si:note:audit_participant_protocol}).
This read-only interface allowed participants to ask questions grounded in the codebase and checkpoints and inspect supporting artifacts (Supplementary Note~\ref*{si:note:audit_ui}).

To score error identification, we use GPT-5.6-Sol at medium reasoning effort to match each participant's original findings to the catalog of 36 applied errors, withholding participant identity, cohort, and historical matching assignments from the judge. 
A match requires identification of the specific altered claim.
We repeat the evaluation five times per participant with identical inputs, yielding 40 independent judgments of 54 findings. 
Match sets agree across all five runs for 53 of 54 findings (98.1\%), with exact agreement in 536 of 540 pairwise comparisons (99.3\%). 
We manually adjudicate the two findings that either vary across runs or disagree with historical assignments, retaining the unanimous matches for all other findings. 
For each participant, we count the reported findings, finding--error matching assignments, and unique inserted errors recovered, deduplicating catalog identifiers across findings for the last measure. Figure~\ref{fig:audit}f reports total findings and unique inserted errors recovered per participant. Supplementary Note~\ref*{si:note:human_audit_scoring} details the matching rubric, robustness analysis, and manual adjudication.

\subsection{Scientist-steered discovery case studies}
\label{sec:methods:case_studies}

We use hypothesis steering for three open-ended case studies in extragalactic astronomy, economics, and functional genomics.
In each study, a scientist supplies the hypothesis for each subsequent round, and \name resumes from the preceding checkpoint. 
We describe the input data, study-specific configurations, and completed runs below; shared settings are given in Methods Sec.~\ref{sec:methods:configurations} and Supplementary Note~\ref*{si:note:model_config}.

\xhdr{Extragalactic astronomy experiment setup} We study low-redshift galaxy selection and photometric-redshift regression using $128\times128$ Legacy Survey image cutouts in the $g$, $r$, and $z$ bands together with catalog photometry and reference redshifts.
The research split contains 181,959 training and 12,130 validation galaxies; the original run defines the low-redshift target as $z<0.03$. 
We seed hypothesis generation with a small collection of scientist-supplied papers and use Claude Opus~4.6 for both the director and execution agent, with director context compaction and early stopping enabled. 
This experiment was driven primarily by a research scientist in astrophysics who has a background in machine learning applied to astronomy but limited background in photometric redshift.
The scientist collaborated with a PhD student in machine learning as well as a postdoctoral researcher with more experience in astronomical catalog analysis.
An initial exploratory round characterizes the data, after which a scientist supplies hypotheses for 15 further rounds. 
We set a ten-query cap per round, increased to twenty in Round~9 due to longer experimentation requirements. 
Across the 16 rounds, validation results guide model development and selection; the final round consolidates standardized predictions for 125 models, of which 123 support classification metrics. 
We subsequently retrain selected recipes on separate external development data and evaluate them on an external test split, with the balanced v4 evaluation using a $z<0.05$ target. 
Supplementary Notes~\ref*{si:note:dwarf_trace} and~\ref*{si:note:dwarf_dataset} describe the trajectory and split definitions, and Supplementary Table~\ref*{si:tab:dwarf_training} gives the training configurations. Supplementary Note~\ref*{si:note:dwarf_single_modality_baselines} gives the architectures and training configurations of the two single-modality baselines.

\xhdr{Economics experiment setup} We begin with an open-ended instruction to identify economically interesting patterns in American Community Survey (ACS) data and an empty data directory. 
We enable the Census data agent to retrieve ACS summary tables, confidentiality-protected Public Use Microdata Sample (PUMS) records, and harmonized IPUMS data on demand, returning data files and variable dictionaries for the execution agent to analyze. 
The study draws on state- and county-level demographic, income, housing, and mobility tables concentrated in 2014, 2023, and 2024; 2023 PUMS records for California, Florida, Louisiana, Minnesota, Ohio, and Texas; 2018 Florida housing records; and a 2023 IPUMS USA householder extract. 
We use Claude Opus~4.6 for the director and execution agent, enable director context compaction, and allow up to 25 tool-agent queries per round to accommodate iterative data retrieval and analysis. 
An economist, who is a PhD student and expert in insurance markets, supplies hypotheses between rounds. 
Supplementary Note~\ref*{si:note:econ_trace} summarizes the hypothesis and outcome of every round.
The run completed seven rounds and 160 queries: ten in the initial round and 25 in each subsequent round.
Supplementary Note~\ref*{si:note:census_data} describes the accessed data; the follow-up analyses below distinguish subsequent reproductions and extensions from this checkpointed run.

\xhdr{Functional genomics experiment setup}
We ask \name to investigate how gene perturbations can inform the interpretation of non-coding variation, using a genome-scale CRISPR interference (CRISPRi) study targeting 12,748 genes in primary human CD4$^{+}$ T cells from four donors \cite{zhu2025genome}. 
We supply a differential-expression object containing 33,983 perturbation--condition profiles across 10,282 measured genes under resting, 8-hour-stimulated, and 48-hour-stimulated conditions. 
We enable the OpenTargets data agent throughout to retrieve curated target--disease associations and related annotations for downstream analysis, and add BioMCP as an execution-agent skill beginning in Round 4. 
The director uses Claude Opus~4.6 throughout; execution is configured with Sonnet 4.5 for Rounds 0--2 and Opus 4.6 thereafter. 
We enable director context compaction and early stopping, with a cap of 25 tool-agent queries per round. 
Following an exploratory first round, a scientist supplies hypotheses for 12 further rounds, including the focus on autoimmune disease that develops during the study. 
The completed run spans 13 rounds. 
Supplementary Notes~\ref*{si:note:gwcd4_rounds} and~\ref*{si:note:gwcd4_relay_standardization} describe the original research trajectory and the subsequent standardized network analyses used for the reported follow-ups.

\section*{Data Availability}

The StarEmbed dataset is available at \url{https://huggingface.co/datasets/StarEmbed/ZTF_40k}, and the ProteinGym datasets are available through \url{https://github.com/OATML-Markslab/ProteinGym\#resources}. We provide the original \name implementation and campaign configurations at \url{https://github.com/owencqueen/DiscoveryAgent-code} and an accessible application to try our DiscoveryAgent at \url{https://github.com/evangorrell/DiscoveryAgent-app}. Original agent-generated analysis code, selected outputs, and reproduction instructions for the extragalactic astronomy, economics, and functional genomics case studies are available at \url{https://github.com/owencqueen/DiscoveryAgent-reproduction}. The reproduction repository references the remaining data sources and documents how to obtain the inputs required for these analyses.

\enlargethispage{\baselineskip}
\section*{Acknowledgements}

We thank Sam Alber, Daniel Lee, Rahul Thapa, Andrew Shen, Peter Eckmann, Valentina Giunchiglia, Daniel Chang, Nitya Thakkar, Marcelo Alvarez, Jake Silberg, and other members of the Zou lab for helpful discussion and assistance on the project. O.Q., C.Y., and H.G.Z. are generously supported by graduate fellowship awards from Knight-Hennessy Scholars at Stanford University. H.G.Z. is also supported by the Stanford University Medical Scientist Training Program (NIH T32-GM145402). S.D. is generously supported by the Stanford Interdisciplinary Graduate Fellowship and Stanford HAI Data Science Scholars Program at Stanford University. C.Y. is also supported by the Stanford EDGE Doctoral Fellowship and the National Science Foundation Graduate Research Fellowship Program under Grant No. DGE-2146755. Any opinions, findings, and conclusions or recommendations expressed in this material are those of the author(s) and do not necessarily reflect the views of the National Science Foundation. E.G. acknowledges the University of Tennessee Office of Research for funding support.

\section*{References}

{
\spacing{0.85}

}

\end{document}